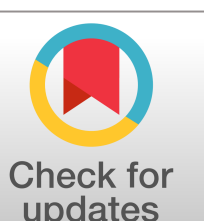

# Deep Transformer Network for Monocular Pose Estimation of Shipborne Unmanned Aerial Vehicle

Maneesha Wickramasuriya,* Taeyoung Lee,† and Murray Snyder‡
*The George Washington University, Washington, D.C. 20052*

**This paper introduces a deep transformer network for estimating the relative six-dimensional (6D) pose of an unmanned aerial vehicle (UAV) with respect to a ship using monocular images. A synthetic data set of ship images is created and annotated with two-dimensional keypoints of multiple ship parts. A transformer neural network model is trained to detect these keypoints and estimate the 6D pose of each part. The estimates are integrated using Bayesian fusion. The model is tested on synthetic data and in situ flight experiments, demonstrating robustness and accuracy in various lighting conditions. The position estimation error is approximately 0.8 and 1.0% of the distance to the ship for the synthetic data and the flight experiments, respectively. The method has potential applications for ship-based autonomous UAV landing and navigation.**

## I. Introduction

UNMANNED aerial vehicles (UAVs) have seen a surge in usage across a multitude of industries, such as aerial photography, military operations, agriculture, mapping, and surveying. The advantages of UAVs over traditional manned aircraft are numerous, including cost-effectiveness, enhanced safety, and superior flexibility. However, the autonomous operation of UAVs, particularly their ability to land on moving platforms like ships, poses a crucial challenge [1]. The unique characteristics of maritime environments (constant motion, unpredictable wave patterns, and rapidly changing atmospheric and light conditions) significantly amplify the complexity of autonomous landing. This challenge is particularly critical in maritime and naval operations, where precision and adaptability are paramount. This capability is of significant importance for industries that depend on maritime transportation or offshore operations.

A primary challenge in this context is the estimation of the UAV's relative pose with respect to the ship, which is vital for precise control of the UAV's movements and ensuring a safe landing. Conventionally, the relative pose has been determined using the Real-Time Kinematic (RTK) GPS. The RTK GPS provides highly accurate positioning by using differential corrections to account for errors in GPS signals. To receive RTK GPS, a communication link between the ship and the UAV must be maintained at all times, typically via radio [2]. Consequently, the UAV relies on the additional measurements on the ship, which limit its ability to independently determine its relative position. In military surveillance operations, the broadcasting of radio signals to maintain this communication link could potentially reveal the ship's position, thereby increasing its vulnerability. Also, reliance on GPS can lead to issues in situations where GPS signals may be weak or unavailable, and it is susceptible to malicious activities, such as jamming or spoofing, which can disrupt the operation of the UAV.

*Ph.D. Candidate, Department of Mechanical and Aerospace Engineering, 800 22nd Street Northwest; maneesh@gwu.edu.

†Professor, Department of Mechanical and Aerospace Engineering, 800 22nd Street Northwest; tylee@gwu.edu (Corresponding Author).

‡Professor Emeritus, Department of Mechanical and Aerospace Engineering, 800 22nd Street Northwest; snydermr@gwu.edu.

An alternative to GPS for estimating the relative pose of a UAV could involve the use of a monocular camera mounted on the UAV, where the captured images can be processed to estimate the UAV's relative pose. Detecting fiducial markers such as lights, laser pointers, or specific visual patterns using cameras are well understood in computer vision, and they are often used for position and attitude estimation during landing operations [3,4]. For example, ArUco markers are composed of square patterns with specific binary codes that can be easily detected and identified by a computer [5].

They have been used to estimate the UAV's relative pose with respect to the ship [6,7]. However, the use of fiducial markers comes with certain disadvantages, such as the need for clear visibility and specific lighting conditions, their susceptibility to occlusion, and the requirement for the markers to be within the camera's field of view. These limitations can affect the accuracy and reliability of pose estimation.

In our preceding studies, we accomplished autonomous flight for shipboard launch and landing using feature-based visual-inertial navigation [8]. However, it turned out that the accuracy of estimating the relative pose depends on the proximity of the camera to the landing pad. The quality and quantity of features significantly influence the accuracy of the relative pose. Moreover, the number of features extracted under variable lighting conditions can fluctuate, and features extracted from the environment outside the ship, such as the sky and ocean, adversely affect the accuracy of relative pose estimation.

Recent advancements in object pose estimation with deep learning, particularly the use of convolutional neural networks (CNNs), have shown significant progress. Multiple convolution layers for feature extraction, object detection, and/or segmentation are used for object silhouette prediction to estimate the six-dimensional (6D) pose of an object [9,10]. Recently, a transformer architecture based on the self-attention mechanism, originally introduced for natural language processing, has been successfully applied to various computer vision tasks. The Detection Transformer (DETR) is one such transformer-based neural network architecture for object detection [11]. This has been extended with the YOLOPose architecture [12] to perform multi-object 6D pose estimation using both direct regression and keypoint regression.

In one of our recent studies, we introduced a Transformer Neural Network for Single Object (TNN-SO) model to estimate the 6D pose from the ship [13], where we took the advantage of tracking a part of a ship near the landing pad as a single object rather than tracking the entire ship. However, despite the robustness of the TNN-SO model for variable lighting and the higher accuracy closer to the landing pad, the accuracy of pose estimation significantly decreased for longer ranges approximately greater than 6 m from the center of the landing pad. This is because the visibility diminished as the UAV flies away from the ship.

To address this issue, in this paper, we propose an innovative method that facilitates multi-object detection using the TNN as shown in Fig. 1. In this approach, the ship is decomposed into multiple parts with varying sizes and locations, as opposed to the previous single object detection method. We model a Transformer Neural Network for Multi Object (TNN-MO) by using the DETR architecture to estimate the 6D pose of a UAV relative to the selected parts of the ship. The multiple pose estimates with respect to those parts are integrated in a Bayesian fashion.[§]

More specifically, the proposed TNN-MO is implemented as follows. The performance of deep neural networks is constrained by the quality and quantity of data used in the training and validation steps. However, the absence of a large labeled data set containing dynamically changing environments for a real-world ship poses a significant challenge in effectively training the transformer neural network model, especially when there are multiple objects to be detected. To tackle the challenges in constructing a rich data set in the real world, we generate synthetic images by rendering a three-dimensional (3D) model of the ship with randomly distributed camera poses under various textured environments and backgrounds to train the TNN-MO model. After training the TNN-MO model, it is able to detect keypoints of multiple parts of the ship.

To estimate the 6D poses, we use the Efficient Point-n-Perspective (EPnP) algorithm [14], leveraging the two- to three-dimensional correspondences of the predicted two-dimensional (2D) keypoints of multiple parts of the ship and its known 3D keypoints. Then, the resulting pose estimates of multiple ship parts are filtered when the object class confidence is greater than 0.9, and the filtered pose estimates are integrated using Bayesian fusion. Upon training both the TNN-SO [13] and TNN-MO models, we observed that the position estimation accuracy of the TNN-MO model surpassed that of the TNN-SO model, even at extended ranges, when evaluated with the synthetic test data set.

To assess the performance of our models in real-world scenarios, we collected images from the camera-attached data collection system (DCS) [15] on an octocopter UAV launched from the YP689 vessel, operated by the United States Naval Academy (USNA) [16]. The DCS is also equipped with RTK GPS, along with a base station mounted at the flight deck of the ship and a dedicated wireless communication link, to measure the relative position up to a centimeter-level accuracy, which is considered as a reference. It is verified that the position error of the proposed TNN-MO in the real world is approximately 2% of the range.

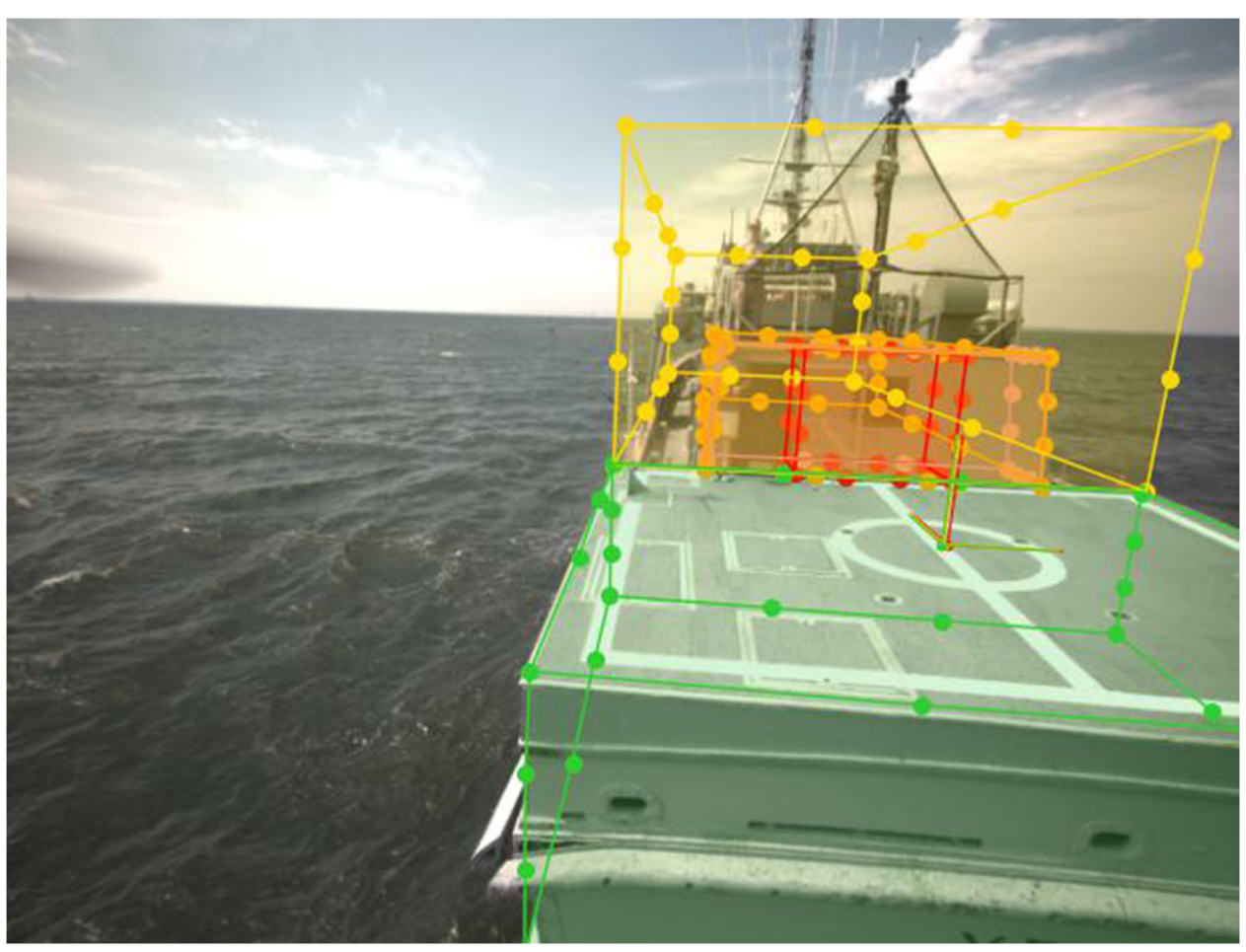

**Fig. 1 We propose a TNN-MO for 6D pose estimation from RGB image, which exhibits excellent accuracy and shows promising potential for vision-based UAV landing.**

[§]Source code: https://github.com/fdcl-gwu/TNN-MO, and video is available online at https://youtu.be/ZG_zVVS8xw8 [retrieved 18 June 2025].

Our proposed approach successfully estimates the relative pose with monocular real-world images, even under challenging conditions such as varying light and different viewpoints. Notably, the TNN-MO model, compared to the TNN-SO model, demonstrates remarkable pose estimation accuracy over a longer range by leveraging the benefits of multi-object detection, while maintaining high accuracy near the landing area. This demonstrates promising potential for vision-based UAV landing and navigation by eliminating the need for GPS, highlighting the effectiveness and precision enhancement offered by our method.

This paper is organized as follows. We present a virtual environment for a ship model in Sec. II, from which synthetic data for training and validation are generated in Sec. III. Next, Sec. IV presents the training procedure and the Bayesian fusion scheme, followed by validations with the synthetic data and with flight experiments, respectively, in Secs. V and VI.

## II. Virtual Environments for Synthetic Data

This paper focuses on a specific scenario of pose estimation relative to a research vessel operated by the United States Naval Academy, specifically YP689 (or identical prior research vessel YP700), over Chesapeake Bay, Maryland. However, it is impractical to obtain a large data set of real-world images for a vessel in the ocean with the exact pose under dynamically changing environments, due to the substantial time and cost required. To overcome this, we generate a synthetic data set from a 3D CAD model using open-source rendering software, with which the proposed network is trained. This section outlines the detailed steps involved in creating the virtual environments, including a CAD model of the ship, texture, and illumination conditions.

### A. Development of 3D CAD Model

First, we create a CAD model of the ship by capturing point clouds of the vessel as shown Fig. 2. We captured a footage of the entire ship using a monocular Red, Green, and Blue (RGB) camera mounted on an octocopter [15,16]. Using this video, we generated a 3D point cloud of the ship using structure from motion techniques in computer vision [17]. The resulting 3D point cloud is corrupted by noise, and it incomplete as the octocopter is not allowed to fly near the bow side of the ship. It first was manually filtered to remove outliers; then, the cleaned point cloud was imported into an open-source rendering software, namely, Blender, to create a CAD model [18]. During this step, detailed structures on the ship not captured by the point cloud, such as the ladders, ship bow, and fence, were manually added too. Furthermore, the CAD model was rescaled according to the actual size of the ship. To enhance the realistic view of the scene, we included other objects such as the ocean, sun, and humans. Also, to make ocean more realistic, we randomly generated wave patterns for every scene using Blender wave modifier tool.

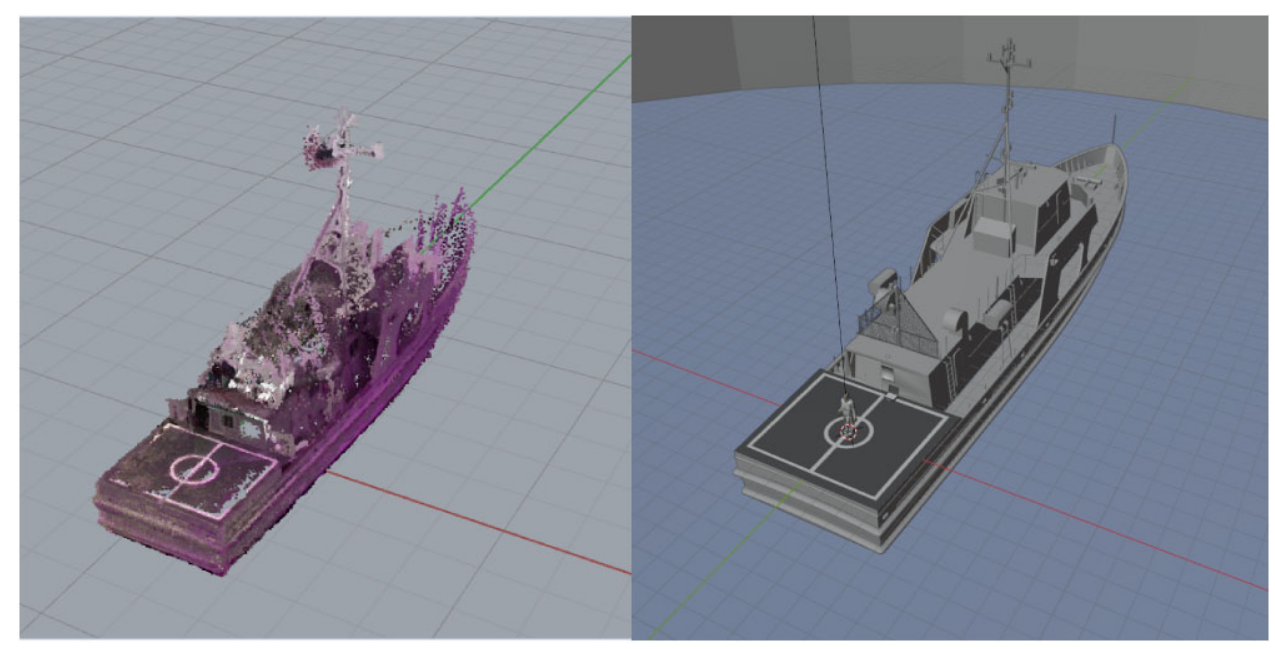

**Fig. 2 3D model based on the filtered point cloud and a CAD model obtained from the point cloud.**

### B. Assignment of Textures

When operating a UAV in the vicinity of a ship, the environment is subject to frequent and dynamic changes. Real images are often influenced by variations in lighting, weather conditions, changes in cloud and wave patterns, occlusions, and other factors. These variables make the detection, recognition, and pose estimation of vessels a challenging task. Training a neural network using synthetic images generated in a static environment may not yield successful results due to the discrepancy between the simulated and real environments, a phenomenon referred to as the sim-to-real gap.

To address this issue, we generate synthetic images using domain randomization, where nonphotorealistic and realistic textures are randomly assigned to each object in Blender [19]. Texture assignment, which involves mapping necessary textures onto 3D objects, is a crucial step in the process of generating synthetic images. In this work, we used two sources to create texture categories: the Describable Textures Dataset [20] and the open-source platform OpenGameArt [21]. The categorization of these textures was based on their image frequency and the characteristics of the textures. Specifically, textures with high frequencies or those resembling oceans were grouped together as *oceanlike textures*. Textures selected with low frequencies were categorized as *vessel skin textures*. Similarly, textures with low frequencies or those resembling the sky were grouped together as *skylike textures*. To define the landing pad and its markings, we grouped together dark, asphaltlike textures as *landing pad textures*, while whitish color textures were categorized as *landing pad markings textures*, as shown in Fig. 3.

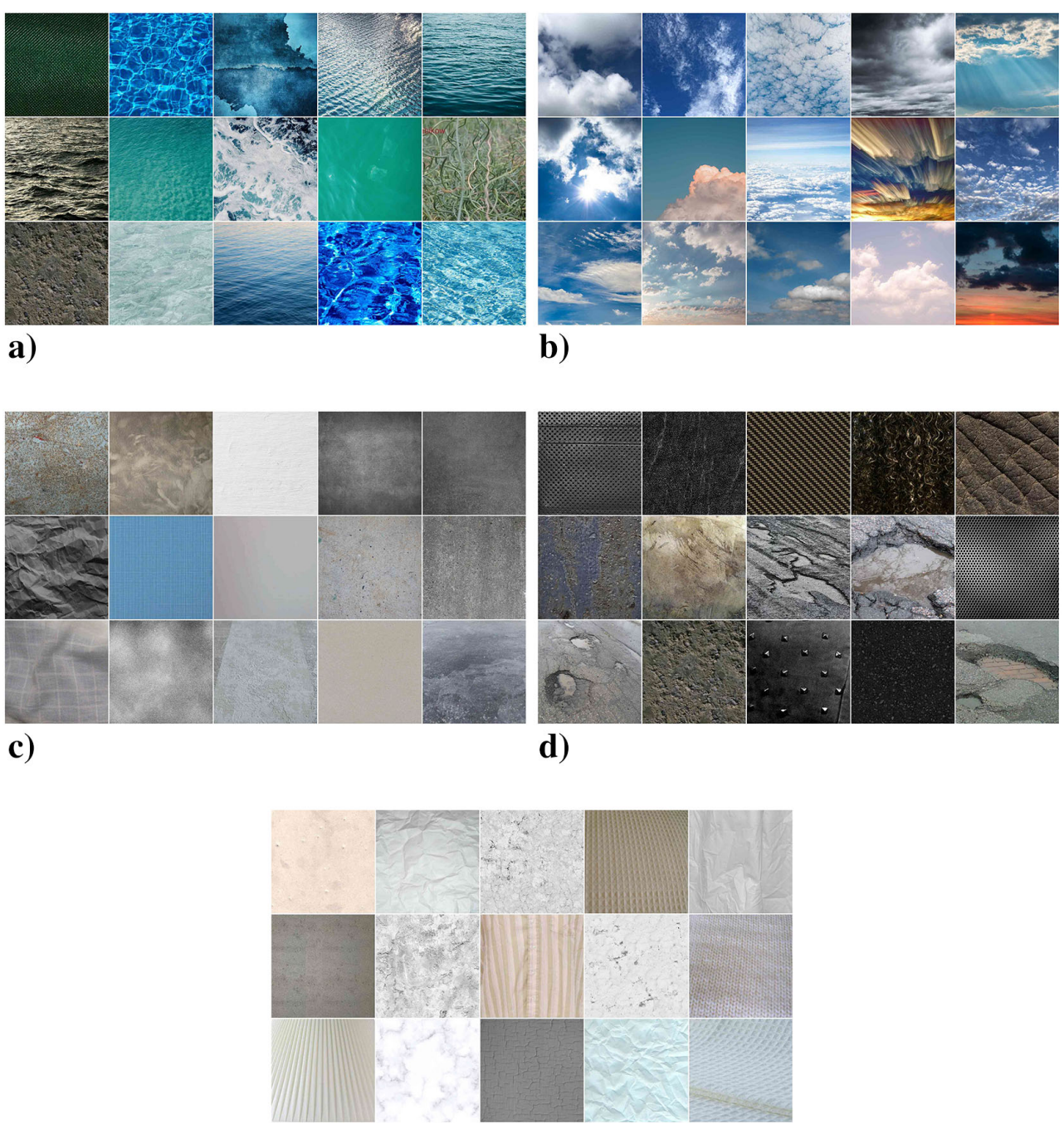

**Fig. 3 Texture categories: a) oceanlike, b) skylike, c) vessel skin, d) landing pad, and e) landing pad markings. These texture categories are used for texture assignment in our study.**

Randomly assigning different textures to objects can enhance object detectability in dynamic environments, which is beneficial for improving the performance of the neural network (NN) model, especially in real-world environments, thereby bridging the sim-to-real gap. Texture categorization is also important as it allows the NN to recognize the shape of the 3D object (vessel) instead of solely relying on object features.

### C. Environmental Illumination and Occlusions

In the ocean environments, the lighting conditions may vary drastically. Furthermore, the illumination of objects in an image can be affected by changes in the orientation of the objects or the camera, even under consistent lighting conditions. For instance, the orientation of an object or the camera may result in the appearance of shadows under direct sunlight. Most cameras are equipped with an automatic exposure control (AE) system that adjusts the brightness of the image based on the lighting conditions. As a result, the lighting conditions in the captured image can vary depending on the adjustments made by the camera's built-in AE system. This variability in lighting conditions poses challenges for object detection, tracking, and pose estimation when using real images. As such, randomizing object texture may not fully mitigate these challenges. To address this, we have enhanced the scene by introducing randomly varying lighting conditions, as illustrated in Fig. 4.

In real-world scenarios, human operators and other objects are present in the vessel, leading to occlusions that cannot be overlooked. For example, during any UAV operation to collect real test data, the operator must be positioned on the landing pad. The occlusions created by such instances become critical during the UAV's landing and takeoff stages. To account for these occlusions, we have incorporated human models and other objects randomly during data generation.

## III. Data Generation in Synthetic Environments

In this section, we incorporate a virtual camera object into the scene to create synthetic images. Importantly, the camera's properties were configured to match those of the actual camera used for image collection, particularly in terms of field of view and resolution. This step is vital in ensuring the accuracy of the pose estimation for real images.

The camera can be positioned anywhere within the 3D space by defining its position and orientation. One of the primary objectives of this work is to estimate the pose of the camera with data-driven learning. Therefore, it is crucial to incorporate various random camera poses in the data set. The distribution of the camera pose chosen for the training data should ideally reflect the distribution of the pose encountered during nominal shipboard operations for launch and recovery. To ensure this, we formulate a probability density function (PDF) for the camera pose. This PDF is designed such that the density value increases near the landing pad when the camera is pointed toward the bow side. This distribution is then dispersed to cover the flight envelope of nominal operations. The training data set is generated by sampling the camera pose from this distribution and generating the corresponding image in the synthetic environment, as discussed in the previous section. The process of sampling the relative pose of the camera involves three steps: defining the distribution of the camera pose, denoted by $C$; defining the distribution the camera is pointed toward, denoted by $F$; and formulating the attitude of the camera based on these distributions [13].

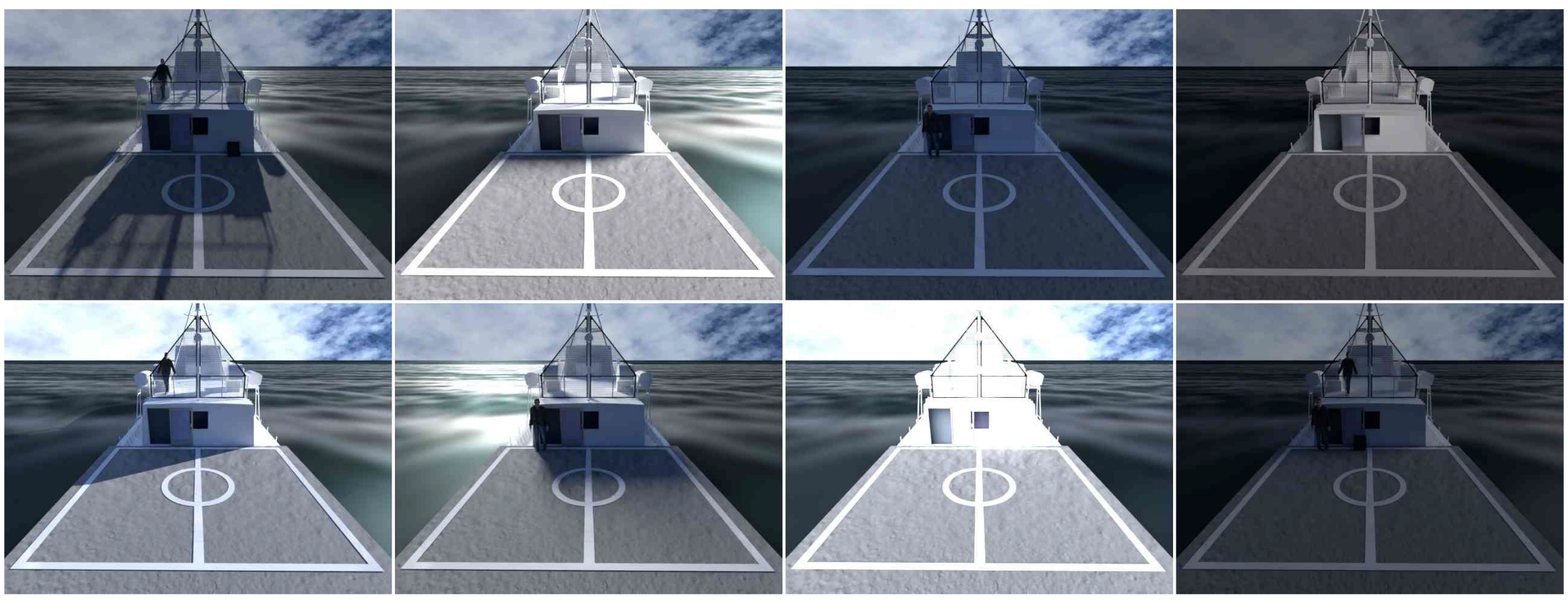

**Fig. 4 Illustration of randomly varied ambient lighting conditions and human models.**

### A. Formulating Probability Density for Position

We first present the probability distribution for $C$ and $F$. Let the base frame $B$ be fixed to the flight deck of the ship. The $x$ axis of this frame points to the starboard side of the ship (right side of the helmsman), the $y$ axis points toward the bow (front) of the ship, and the $z$ axis points upwards. These axes are denoted by $x_B$, $y_B$, and $z_B$, as illustrated in Fig. 5. The location of $C$ or $F$ is defined by the spherical coordinates $(r, \theta_B, \phi_B)$, where the scalar $r \in \mathbb{R}$ corresponds to the distance, and the angles $\theta_B \in [-\pi, \pi]$, and $\phi_B \in [0, (\pi/2)]$ represent the rotation about the $z$ axis and the rotation about the $x$ axis, respectively.

For the position of the camera $C$, the angle $\theta_B$ is sampled from a uniform distribution on the range $[-\pi, 0]$, representing the area behind the flight deck. Similarly, $\phi_B$ is uniformly sampled from the range $[0, (\pi/3)]$, corresponding to the area above the flight deck. This choice is motivated by our desire to generate realistic camera orientations that resemble those of a horizontally fixed camera during UAV flight. We initially experimented with increasing the maximum range of $\phi_B$ closer to $\pi/2$ but found that this led to a number of unrealistic camera orientations. To mitigate this, we decided to set the maximum range for $\phi_B$ at $\pi/3$. This decision has proven effective in maintaining the realism of our generated camera orientations. The distance $r$ is sampled from a truncated normal distribution, which is a one-dimensional Gaussian distribution truncated to the range of $[0, L]$ and rescaled accordingly. Here, the maximum range $L$ is chosen as 25 m, and the mean and the standard deviation of the distribution are 1 and 40, respectively. Next, the distribution for the target point $F$ is sampled in a similar manner. The ranges for each of $\theta_B$, $\phi_B$, and $r$ are $[0, 2\pi]$, $[-(\pi/2), (\pi/2)]$, and $[0, 15]$, respectively. The mean and the standard deviation for the truncated normal distribution are 0 and 1. The resulting set of sampled $C$ and $F$ is presented at Fig. 6.

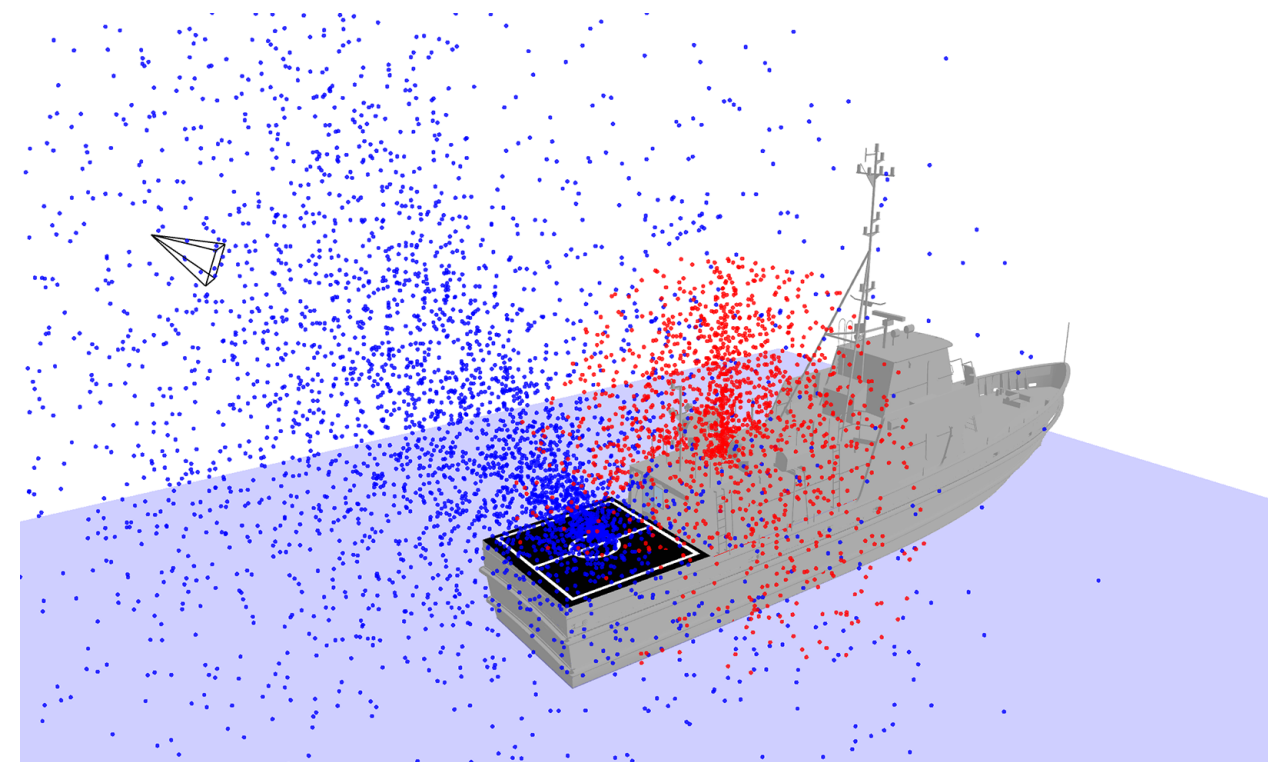

**Fig. 6 Sampled points for the camera location $C$ (blue) and the points that the camera is pointed toward $F$ (red).**

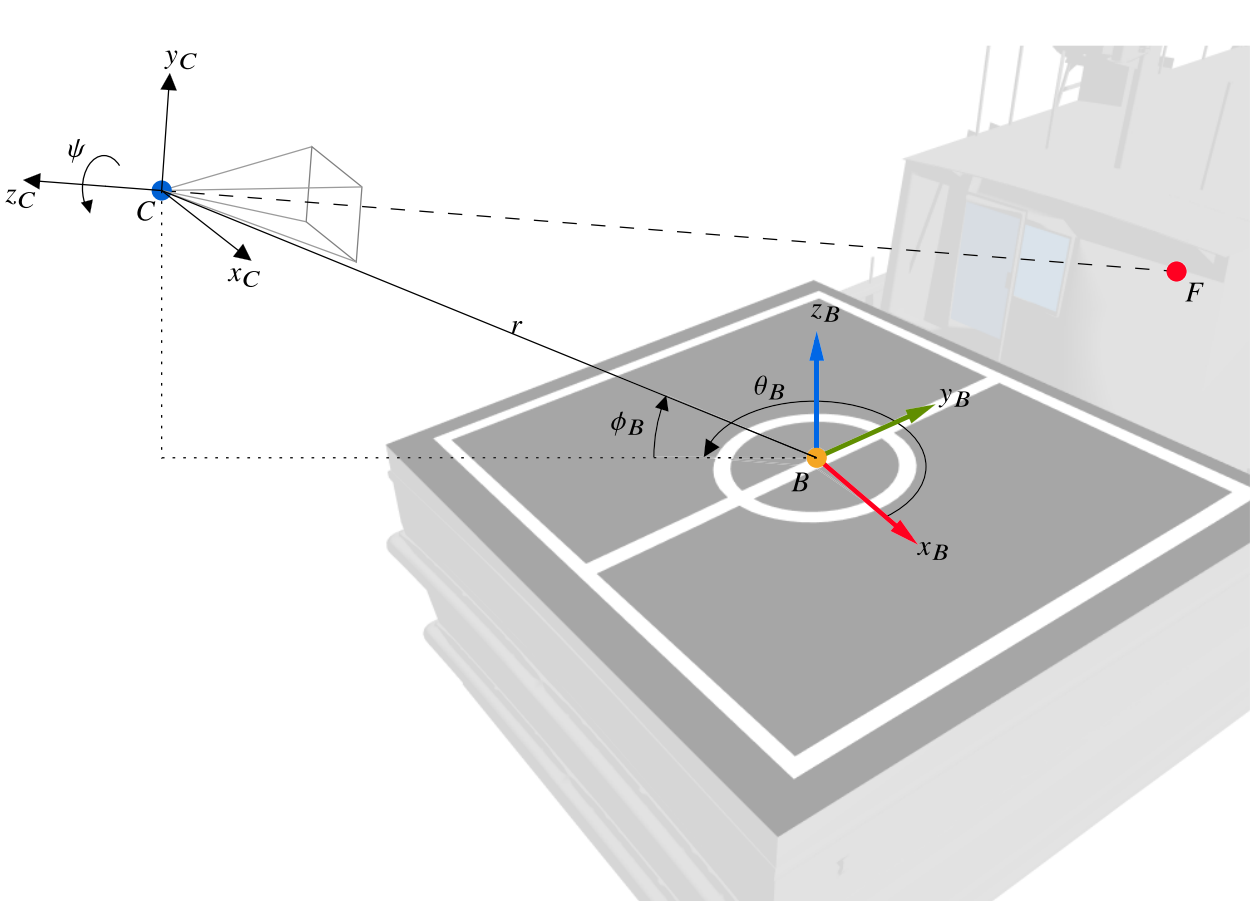

**Fig. 5 The base frame and the camera frame.**

### B. Attitude of Camera

Once the points $C$ and $F$ are sampled, the camera attitude is determined as follows. Let the camera frame be defined such that its origin coincides with $C$, and the positive $x$ axis points to the right (along the direction of the image width), the positive $y$ axis points upward (along the direction of the image height), and the positive $z$ axis is opposite to the line of sight, as illustrated in Fig. 5. The orientation of the camera frame with respect to the flight deck frame is specified by a rotation matrix $R \in \mathsf{SO}(3) = \{R \in \mathbb{R}^{3\times3} | R^T R = I, \det[R] = 1\}$.

The points of $C$ and $F$ are sampled as described previously, and they are reordered randomly to create a set of pairs of $(C, F)$. Each pair defines the line of sight of the camera as $\boldsymbol{CF}$. We choose the corresponding rotation matrix $R' \in \mathsf{SO}(3)$ such that its third axis is opposite to $\boldsymbol{CF}$ and its $x$ axis is level to the ground. More specifically, $R'$ is given by

$$R' = [r_1', r_2', r_3']$$

where the unit vectors $r_1', r_2', r_3' \in \mathbb{R}^3$ are obtained by

$$r_3' = -\frac{\boldsymbol{CF}}{\|\boldsymbol{CF}\|}, \quad r_1' = \frac{e_3 \times r_3'}{\|e_3 \times r_3'\|}, \quad r_2' = r_3' \times r_1'$$

with $e_3 = [0, 0, 1] \in \mathbb{R}^3$. Then, the camera attitude is obtained by rotating $R'$ about the $z$ axis by an angle $\psi$, which is sampled from the uniform distribution in the range $[-(\pi/6), (\pi/6)]$, that is, $R = R' \exp(\psi \hat{e}_3)$, where the hat map $\hat{\cdot}: \mathbb{R}^3 \times \mathbb{R}^{3\times3}$ is defined such that $\hat{x}y = x \times y$ and $\hat{x} = -\hat{x}^T$ for any $x, y \in \mathbb{R}^3$. The angle $\psi$ represents the possible rolling motion of the camera.

### C. Generated Synthetic Images

The selected camera pose is transferred to the rendering software Blender to generate the synthetic images with varying textures, wave patterns, and lightening conditions. Each image is paired with the correct pose $(C, R)$ to be used for training and verification. The selected samples from the synthetic images are illustrated in Fig. 7.

In the virtual environment, the sea is modeled as an infinitely large plane, leading to the prominence of the sea horizon (the line where the sea meets the sky) in the background of most synthetic images. However, in real-world scenarios, the visibility of the sea horizon can be obstructed by various objects or geographical features. For instance, when the ship is in a bay, the shoreline or the land may alter the view of the horizon. As such, a data set with infinitely extending sea may not represent the real-world scenario properly. We have observed that a network trained over such a data

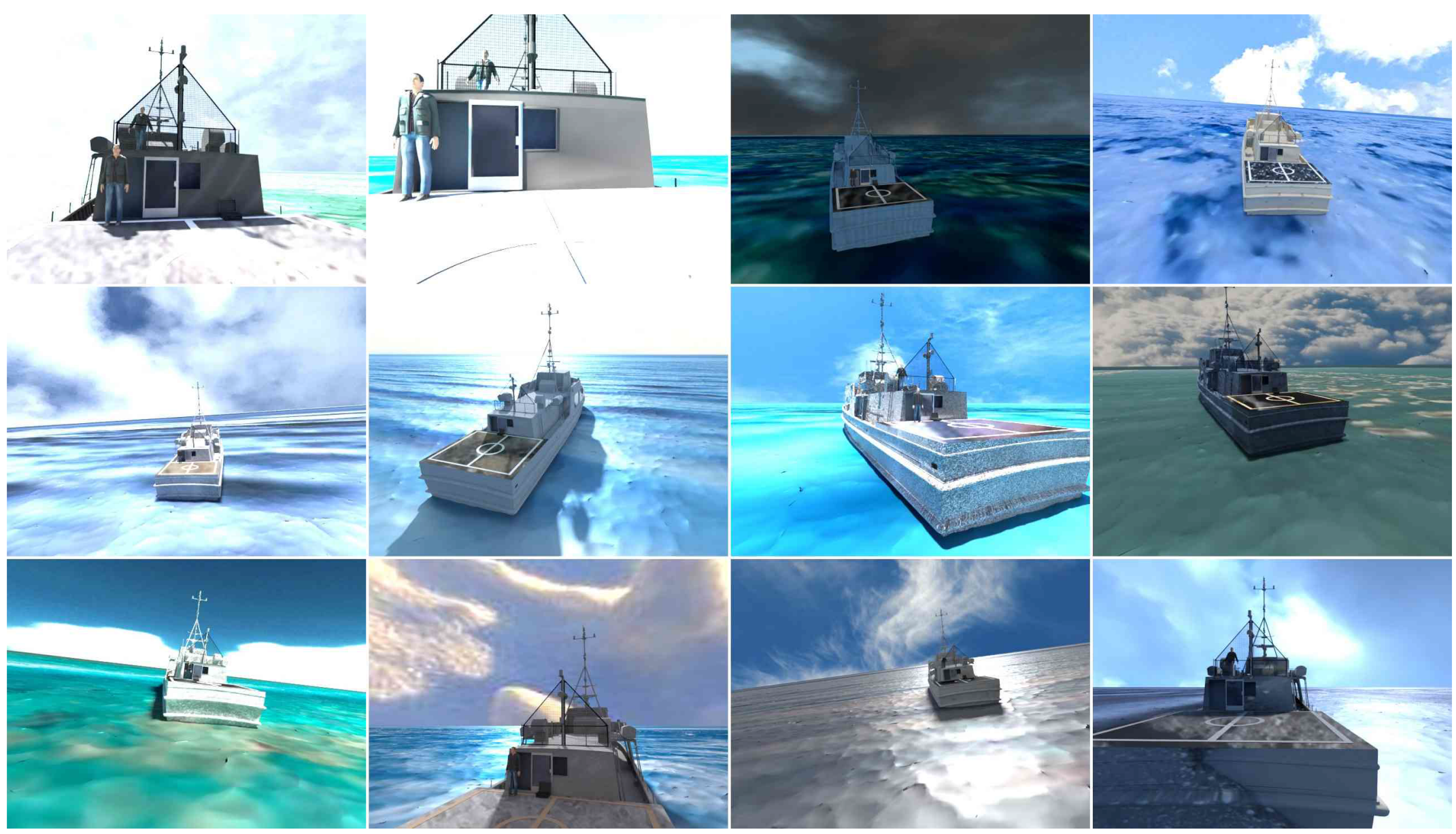

**Fig. 7 Generated synthetic images with the presence of sea horizon.**

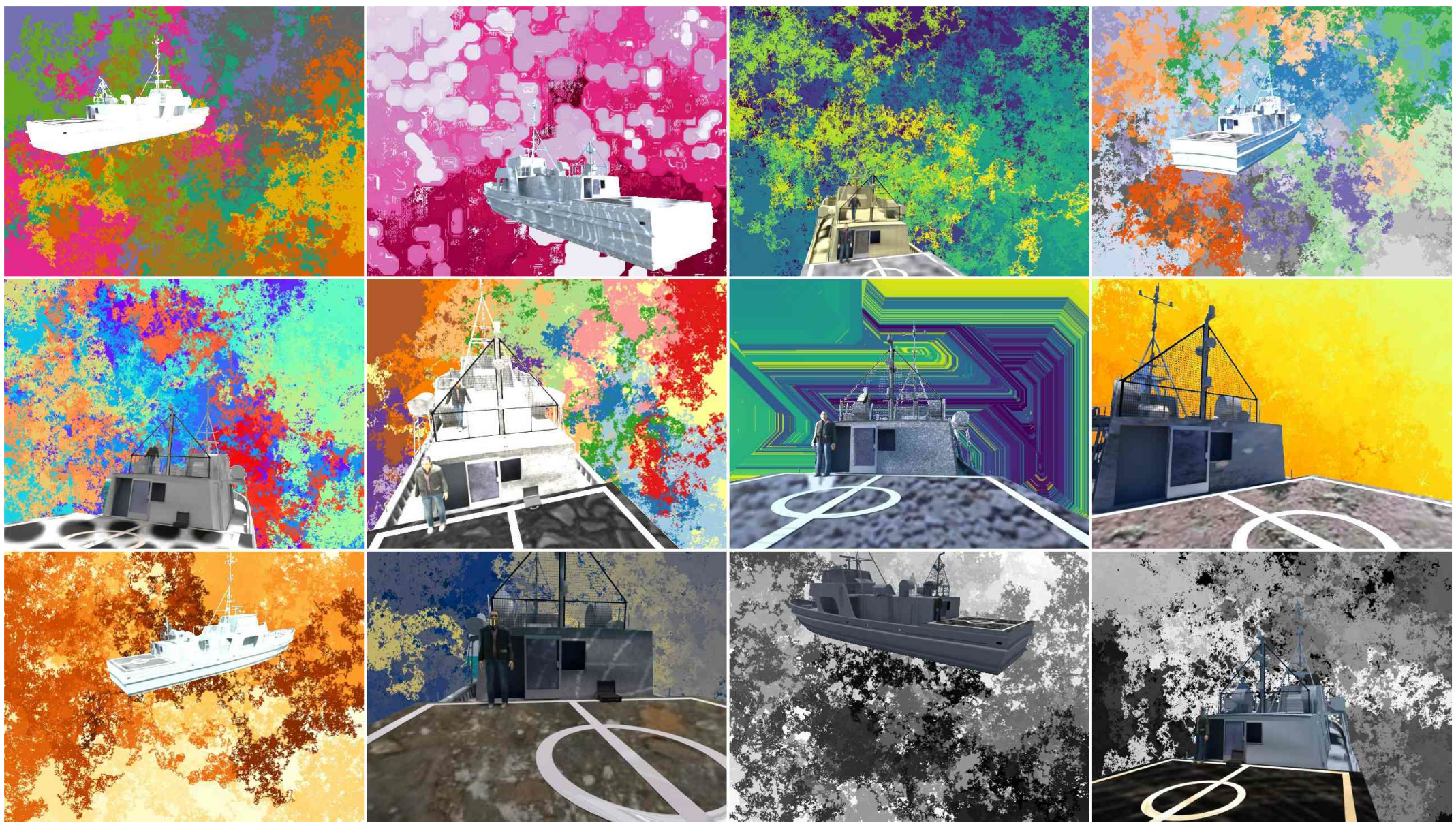

**Fig. 8 Generated synthetic images without sea horizon by replacing background with random image.**

set may exhibit a bias especially in estimating the height of the camera.

To address this, we have opted to generate a distinct synthetic image data set without the horizon. This is achieved by substituting the ship's background with a randomly generated image. The random image is produced using the random image generator as proposed by Ref. [22]. Then, we mix the mentioned data set with the synthetic images without any horizon as shown in Fig. 8. This approach ensures a more diverse and realistic representation in our synthetic image data set, while capturing the importation information provided by a horizon.

### D. Decomposition of Ship into Multiple Parts and Keypoint Selection

The synthetic images as shown in Figs. 7 and 8 serve as the input to the neural network formulated in the next section. The desired outputs, or the target data correspond to the pixel location in the image for the 3D keypoints defining a predefined bounding box of the ship. Then, the camera pose can be estimated by the resulting correspondence between the 2D image coordinate and the 3D location of the keypoints.

However, defining the entire ship as a single object is not desirable, as the visibility might be obscured depending on the position and the orientation of the camera. For example, during the landing, the keypoints at the stern of the ship may not be visible. To address this issue, the ship is decomposed into multiple parts for detection, rather than being detected as a whole. This method also helps mitigate the impact of certain camera orientations and specific lighting conditions that might result in only partial visibility of certain areas of the ship, potentially affecting the accuracy of the model's estimations. Specifically, the ship is decomposed into six parts: the whole ship, stern, superstructure, and three parts of the flight structure immediately forward of the flight deck, referred to as the dog house, as illustrated at Fig. 9.

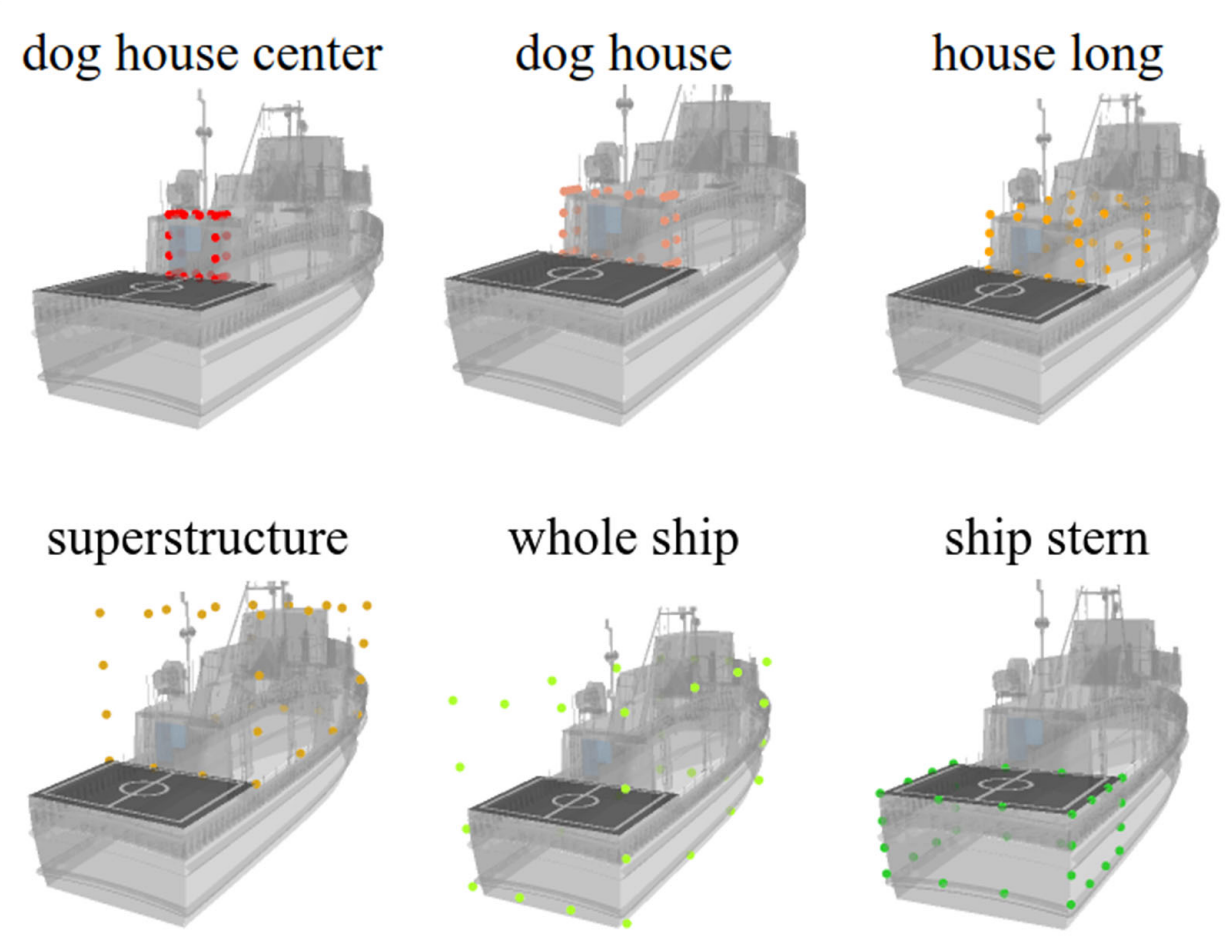

**Fig. 9 Selected specific components of the ship and 3D keypoints.**

For each of these six parts, we define a 3D bounding box by specifying eight corners following Ref. [12]. Each of 12 edges of the bounding box is divided into three sections by adding two intermediate points. As such, a bounding box is defined by 32 keypoints. The location of each keypoint is calculated by the perspective projection from the camera model. In short, the target for each synthetic image is the 2D location of 32 keypoints for each of six parts of the ship.

## IV. Relative 6D Object Pose Estimation

In this section, we present a deep neural network model for pose estimation and a Bayesian fusion to integrate the estimate pose relative to multiple parts of the ship. For the neural network architecture, we employ a TNN as presented in Ref. [12]. In this method, we estimate the 2D keypoints associated with each object and the object class confidence of the RGB image. Then, we recover the corresponding relative 6D poses of each object by solving the 2D to 3D correspondence of the keypoints using EPnP [14]. Finally, we integrate the resulting pose estimations using Bayesian fusion to obtain the most probable pose estimate.

### A. Transformer Neural Network for Keypoints Estimation

The proposed transformer neural network architecture, as shown in Fig. 10, is described as follows. First, for a given RGB image, we extract image features using a CNN backbone. Specifically, the input RGB image has a size of $H \times W = 480 \times 640$. We employ ResNet50 to generate a lower-resolution feature $f \in \mathbb{R}^{C \times H_0 \times W_0}$, where $C = 2048$ is the channel dimension and $H_0, W_0$ are defined as $(H/32) = 15, (W/32) = 20$, respectively, following the DETR [11]. Then, we use a $1 \times 1$ convolution to reduce the feature dimension to $d = 256$. The extracted features, which are in the dimension of $d \times H_0 \times W_0$, are flattened to a dimension of $d \times H_0 W_0$. Subsequently, a fixed positional encoding is added to the features to generate input embeddings, which have the dimension of $d \times H_0 W_0$. Then, each of these embeddings is used as an input for the transformer encoder [23].

The transformer encoder has six standard encoder layers with skip connections. The output from the encoder is passed to the decoder module, which also has six standard decoder layers with skip connections, accompanied by $Q$ object queries [12]. These object queries are a set of learnable positional embeddings with the dimension $d \times Q$, representing the position of each object class. These vectors are learned during the training process of the model, to help the model understand and represent the positional relationships between different objects in the data. Let $N$ be the cardinality of the object classes including the case when no object is detected, which is denoted by $\emptyset$. Because we are considering six parts of the ship as objects which are defined in Fig. 9, we have $N = 7$. In the presented TNN-MO model, we set $Q = N$, implying that the number of object queries is equal to the object classes.

The resulting decoder output embeddings are processed with $N$ feedforward networks (FFNs). Each FFN is composed of two parts: linear layers with an input size of 256 and an output size of $N$ for class prediction and a standard three-layer perceptron that has an input size of 256, a hidden layer dimension of 256 with the Rectified Linear Unit (ReLU) activation, and an output size of 64 for keypoint prediction.

These outputs of the $i$th FFN, corresponding to the $i$th object query, are denoted by the class predition $\bar{c}_i \in \mathbb{R}^N$ and the keypoints $\bar{K}_i \in \mathbb{R}^{32 \times 2}$, respectively. Here, the $j$th variable of $\bar{c}_i$, namely, $\bar{c}_{ij}$, corresponds to the class logit (unnormalized score) that the keypoint output of the $i$th FFN $\bar{K}_i$ belongs to the $j$th object class. In other words, the higher the value of $\bar{c}_{ij}$ is, the more likely it is that the keypoints $\bar{K}_i$ are of the $j$th object. These logits are passed through a softmax function to obtain the predicted class probability $\bar{p}_i \in [0, 1]^N$ of the $i$th FFN, satisfying $\sum_{j=1}^{N} \bar{p}_{ij} = 1$. These are concatenated into a stochastic matrix $\bar{P} = [\bar{p}_1^T, \bar{p}_2^T, \ldots, \bar{p}_N^T]^T \in [0, 1]^{N \times N}$, where $\bar{p}_{ij}$ denotes the predicted probability that the $i$th set of keypoints $\bar{K}_i$ belongs to the $j$th object class.

### B. Ground Truth Data

In the training input image, the ground truth object class labels are represented by $\boldsymbol{c}_g = [c_{g_1}, c_{g_2}, \ldots, c_{g_N}] \in \{0, 1\}^N$, where $c_{g_j}$ is a binary variable indicating the presence (or the absence) of the $j$th object. Specifically, if the $j$th object appears in the training input image, we set $c_{g_j} = 1$; otherwise, we set $c_{g_j} = 0$. The $N$th object corresponds to the $\varnothing$ object, and therefore $c_{g_N} = 0$ always for any image in the training data set. It is important to note that, because we are considering a single ship, there is no repetition of objects in the image. For example, following the object classes defined in Fig. 9, when the whole ship (fifth) and the ship stern (sixth) are visible in a specific image, the resulting ground truth object class labels is $\boldsymbol{c}_g = [0, 0, 0, 0, 1, 1, 0]$.

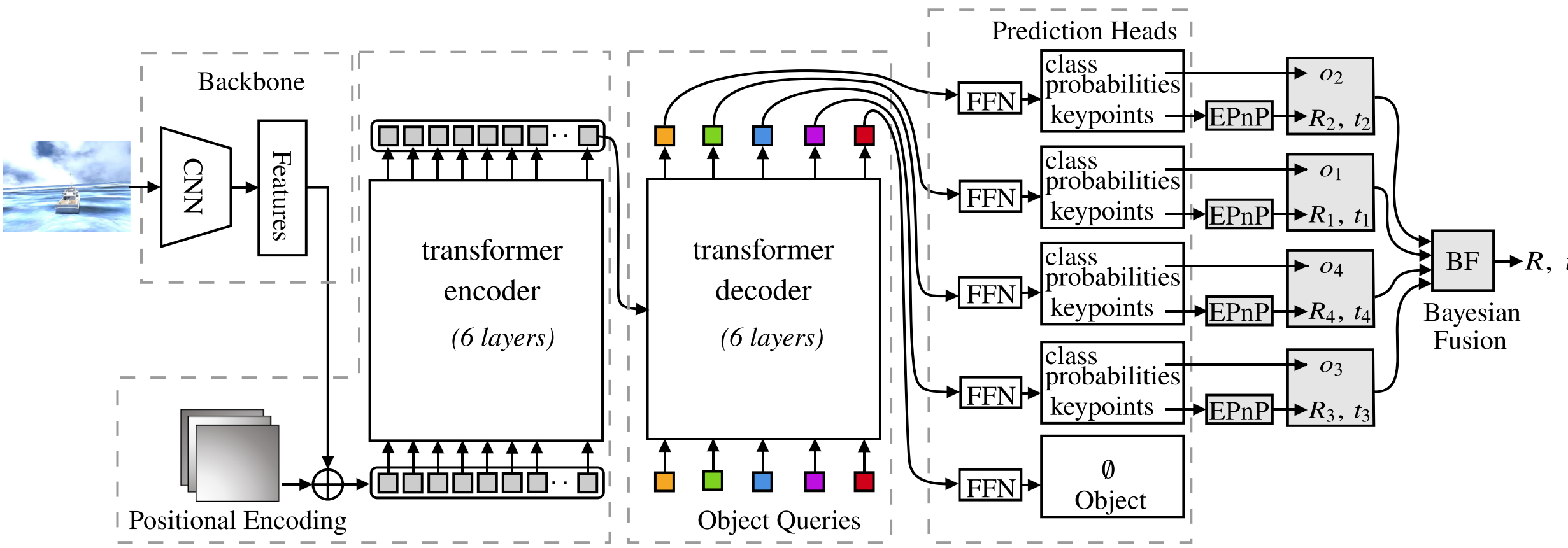

**Fig. 10 TNN-MO model architecture to estimate 6D pose.**

For any $i$ with $c_{g_i} \neq 0$, the corresponding location of the keypoints in the image is denoted by $K_i \in \mathbb{R}^{32\times 2}$, and the location of the keypoints in the ship-fixed frame is denoted by $q \in \mathbb{R}^{32\times 3}$.

### C. Loss Function for Training

For the given labeled data and the predicted output, the loss function for training is formulated as follows. Our approach to loss computation is inspired by the set prediction task, similar to the methods used in DETR [11] and YOLOPose [12]. It is composed of two steps of identifying the matching pairs and calculating the loss.

First, we perform the matching between the predicted class probability outputs and the ground truth using bipartite matching [11]. Let $\mathfrak{S}_N$ be the set of all possible permutations of $\{1, \ldots, N\}$. For any sequence $\sigma \in \mathfrak{S}$, the matching cost is defined by

$$\mathcal{L}_{\text{match}}(\sigma) = \sum_{i=1}^{N} -c_{g_i} \bar{p}_{\sigma(i)i} \tag{1}$$

where $\sigma(i)$ gives the index of the $i$th element in the permuted sequence. As such, this represents the sum of the negative probability that $\bar{K}_{\sigma(i)}$ belongs to the $i$th object class for all objects present in the training image satisfying $c_{g_i} \neq 0$. The bipartite matching is to identify the optimal sequence that minimizes the matching cost,

$$\bar{\sigma} = \underset{\sigma \in \mathfrak{S}_N}{\arg\min}\, \mathcal{L}_{\text{match}}(\sigma) \tag{2}$$

and it is addressed by the Hungarian method [24].

Next, after the matching pairs are identified, we formulate the keypoint loss as the sum of the negative log likelihood for the object class prediction and the error in the prediction of the keypoints, as given by

$$\mathcal{L}_{\text{keypoints}} = \sum_{i=1}^{N} \Big[ -\log \bar{p}_{\bar{\sigma}(i)i} + \gamma c_{g_i} \| K_i - \bar{K}_{\bar{\sigma}(i)} \|_1 \Big] \tag{3}$$

where $\gamma > 0$ is a hyperparameter for relative weighting. The parameters of the proposed deep neural network and the object queries are randomly chosen initially, and they are adjusted to minimize the mentioned loss during training. The detailed implementation of the training is described in the next section.

### D. Model Inference with Bayesian Fusion

During the inference phase, the proposed TNN returns the probability of the class prediction $\bar{p}_i \in [0, 1]^N$ and the predicted keypoints $\bar{K}_i \in \mathbb{R}^{32\times 2}$ for each object query $i \in \{1, \ldots, N\}$. The object class corresponding to the $i$th object query, namely, $\sigma^*(i) \in \{1, \ldots, N\}$ is identified by

$$\sigma^*(i) = \underset{1 \leq j \leq N}{\arg\max}\, \{\bar{p}_{i1}, \ldots \bar{p}_{iN}\}$$

and the resulting maximum probability is denoted by $o_i = \bar{p}_{i\sigma^*(i)} \in [0, 1]$, which is considered as the confidence in the object classification.

Together with the 3D location of the keypoints (or the bounding box), namely, $q_{\sigma^*(i)} \in \mathbb{R}^{32\times 3}$ in the ground truth, the pair $(\bar{K}_i, q_{\sigma^*(i)})$ provides a 2D-to-3D correspondence for 32 points on the $\sigma^*(i)$th bounding box, which can be used to estimate the camera pose with respect to the ship-fixed frame. Specifically, we use the EPnP algorithm [14] in conjunction with RANSAC to estimate the pose $(R_i, t_i) \in \mathsf{SO(3)} \times \mathbb{R}^3$ from the $i$th object.

As such, we obtain at most $N - 1$ pairs of $(R_i, t_i)$ from the object classes defined in Sec. III.D, excluding the predictions corresponding to the no object class. Each pose estimate has a varying degree of accuracy and confidence, as one object may be captured more clearly than others depending on the perspective of the camera and the lighting condition. Or it is possible that certain objects are occluded or outside of the field of view.

To address this, the multiple pose estimates are integrated as follows. It has been empirically observed that the class confidence of the object, namely, $o_i = \bar{p}_{i\sigma^*(i)}$ is closely related to the accuracy of the pose estimate [25]. To eliminate outliers, the estimated pose is discarded if the object class confidence is less than or equal to 0.9, that is, $o_i \leq 0.9$. Let $N_{\text{inliers}} \leq N - 1$ be the number of estimates remaining after removing outliers. Then, the class confidence $\{o_i\}_1^{N_{\text{inliers}}}$ for the inliers are filtered with the softmax function to obtain the normalized weight $w_i \in [0, 1]$ for the inliers, satisfying $\sum_i w_i = 1$. Each pose estimate $(R_i, t_i)$ is paired with the corresponding weight $w_i$, to be integrated into a single pose estimate.

For the position estimate, the mean $\mu_t \in \mathbb{R}^3$ and the covariance matrix $\Sigma_t \in \mathbb{R}^{3\times 3}$ are obtained by the weighted sum as

$$\mu_t = \sum_i w_i t_i, \quad \Sigma_t = \sum_i w_i (t_i - \mu_t)(t_i - \mu_t)^T \tag{4}$$

which serve as the position estimate and the corresponding degree of confidence.

Next, in [26], it has been shown that both the mode and the degree of dispersion on the rotation group can be obtained by the arithmetic mean, or the first moment of the rotation matrix $\mathrm{E}[R] \in \mathbb{R}^{3\times 3}$ computed by

$$\mathrm{E}[R] = \sum_j w_j R_j \tag{5}$$

Note that, while each $R_j$ belongs to $\mathsf{SO(3)}$, the weighted sum is not necessarily on $\mathsf{SO(3)}$. Let $\mathrm{E}[R] = U' D' (V')^T$ be the singular value decomposition of $\mathrm{E}[R]$ with the orthogonal $U', V' \in \mathbb{R}^{3\times 3}$ and the diagonal $D' \in \mathbb{R}^{3\times 3}$. The *proper* singular value decomposition of $\mathrm{E}[R]$ is given by

$$\mathrm{E}[R] = U D V^T \tag{6}$$

where the rotation matrices $U, V \in \mathsf{SO(3)}$ and the diagonal matrix $D = \mathrm{diag}[d_1, d_2, d_3] \in \mathbb{R}^{3\times 3}$ are defined as

$$\begin{aligned} U &= U' \mathrm{diag}[1, 1, \det[U']], \\ D &= D' \mathrm{diag}[1, 1, \det[U'V']], \\ V &= V' \mathrm{diag}[1, 1, \det[V']] \end{aligned}$$

Given the value of $\mathrm{E}[R]$ and its proper singular value decomposition, the maximum likelihood estimate of the attitude is given by $\mu_R = UV^T \in \mathsf{SO(3)}$ [26]. The confidence in the attitude estimate can be measured by the diagonal elements of $D$ satisfying $1 \geq d_1 \geq d_2 \geq |d_3| \geq 0$. Roughly speaking, as the diagonal elements become closer to 1 (respectively, 0), the confidence level on the attitude estimate $\mu_R$ increases (respectively, decreases).

More specifically, the distribution of the estimated attitude can be described by the matrix Fisher distribution, which is the maximum entropy (or most arbitrary) distribution on $\mathsf{SO(3)}$ when conditioned by the fixed value of $\mathrm{E}[R]$. From the matrix Fisher distribution, the degree of uncertainties in the attitude estimate can be quantified as follows. For any $R \in \mathsf{SO(3)}$, let $\eta \in \mathbb{R}^3$ be defined such that $R = U \exp(\hat{\eta}) V^T$. In other words, $\eta$ represents the difference between $R$ and the estimated attitude $\mu_R$ in the sense that $R$ can be obtained by rotating $\mu_R$ about the axis $V\eta$ resolved in the ship-fixed frame (or the axis $U\eta$ when resolved in the estimated camera frame) by the angle $\|\eta\|$. Thus, $V\eta$ is the axis of rotation, and $\|\eta\|$ is the angle of rotation from $\mu_R$ to $R$.

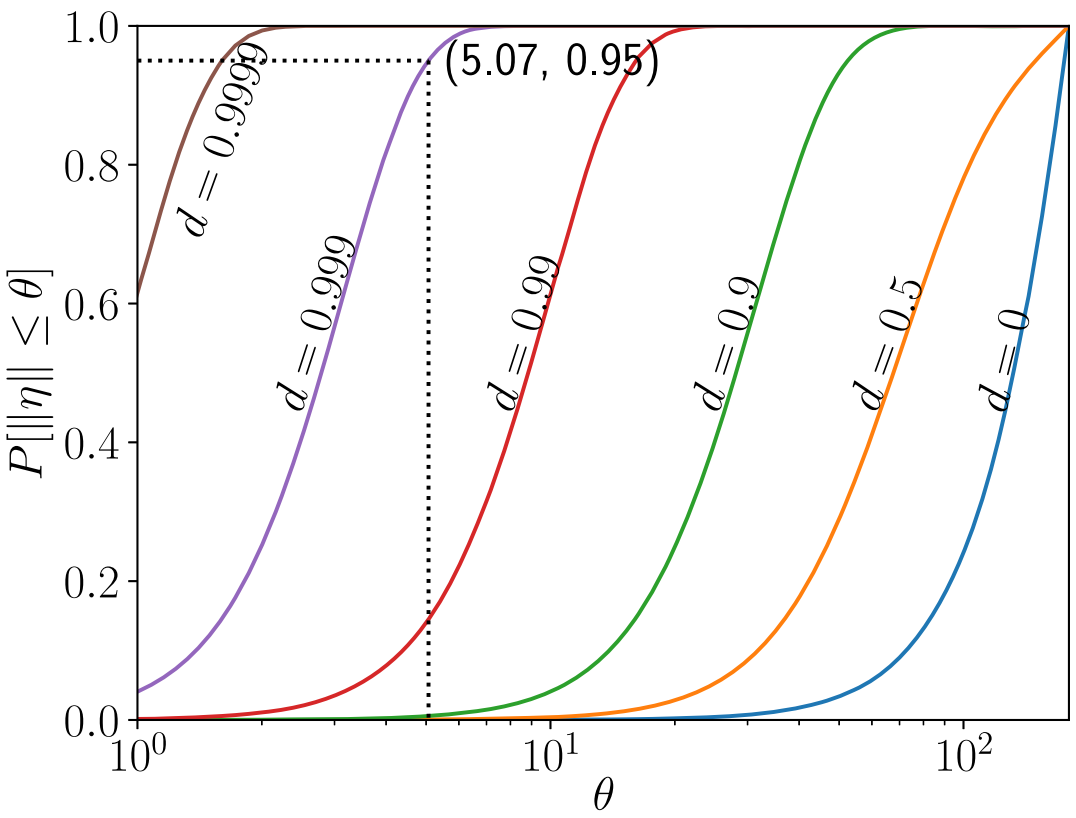

**Fig. 11 Degree of confidence in the attitude estimate.**

There are two methods for uncertainty quantification. First, when the diagonal elements are identical, that is, $d_1 = d_2 = d_3 \triangleq d$, the probability that the angle of rotation is less that a specific value $\theta \in [0, \pi]$ can be computed by

$$\mathbb{P}[\|\eta\| \leq \theta] = \frac{1}{\pi(I_0(2s) - I_1(2s))} \int_0^\theta \exp(2s\cos\rho)(1 - \cos\rho)\, \mathrm{d}\rho \tag{7}$$

where $I_0, I_1$ correspond to the modified Bessel function of the first kind and $s \geq 0$ is a scalar that can be obtained numerically from $d$ [26]. The value of the mentioned probability for varying $d$ and $\theta$ is illustrated at Fig. 11. For example, when $d = 0.999$, the estimation error is less than 5.07° with the probability of 0.95.

Second, when the estimated distribution is highly concentrated, or when $d_3 \to 1$, the rotation vector $\eta$ follows a Gaussian distribution $\mathcal{N}(0, \Sigma_\eta)$ with

$$\Sigma_\eta = \mathrm{diag}[1 + d_1 - d_2 - d_3, 1 - d_1 + d_2 - d_3, 1 - d_1 - d_2 + d_3] \tag{8}$$

As it is unlikely to encounter highly concentrated estimates where Eq. (8) holds in practice, the empirical expression can be used instead,

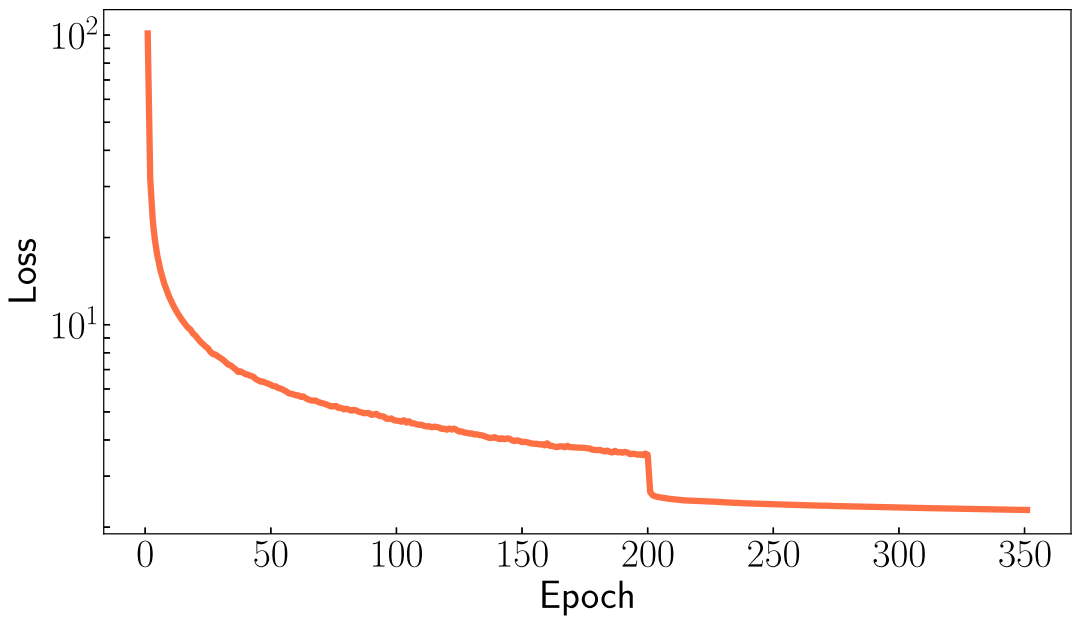

**Fig. 12 Training loss over epochs.**

$$\Sigma_\eta = \sum_i w_i \eta_i \eta_i^T \tag{9}$$

where $\eta_i \in \mathbb{R}^3$ is obtained such that $\hat{\eta}_i = \log(U^T R_i V)$.

In short, the attitude estimate is given by $\mu_R = UV^T$, and the confidence in the estimated attitude can be analyzed with Eq. (7), (8), or (9).

## V. Model Training and Testing with Synthetic Data

The mentioned transformer network and the Bayesian fusion constitute the proposed network architecture to estimate the 6D pose from monocular vision. In this section, we describe how the network is trained, including the implementation details and trained model is validation with the synthetic data.

### A. Model Training with Synthetic Data

The model is trained as follows. The training data set is composed of 435,000 synthetic images, including 332,000 images with a sea horizon background and 102,000 images with a random image background (without a sea horizon). Given that the synthetic data set already included randomized situations, effectively covering a broad spectrum of variations and scenarios, image augmentation techniques were not used during the training process. The existing data set was deemed to provide sufficient diversity and complexity for effective learning and adaptation of the model, rendering additional image augmentation unnecessary.

The TNN-MO model was implemented using the PyTorch framework. The model was trained for 350 epochs with the batch size of 48, hyperparameter $\gamma = 10$, leveraging the AdamW optimizer [27], using a 20 GB multi-Instance GPU partition from NVIDIA A100-PCIE-40GB GPU. The learning rate was initially set to $10^{-4}$ for the first 200 epochs, after which it was reduced to $10^{-5}$. Gradient clipping was also employed, limiting the maximum gradient norm to 0.1. The resulting training curve is presented at Fig. 12.

### B. Model Testing with Synthetic Data

The trained model is validated with the synthetic test data set that was not in the training data set. For further comparison, we also incorporate the TNN model without the proposed multi-objects framework as developed in Ref. [13]. The resulting single object model predicts the relative pose with respect to the *dog house* structure. These are referred to as the TNN-SO Model and TNN-MO Model, respectively.

We compute the mean absolute error (MAE) for the position estimate and the attitude estimate over synthetic images under various lighting conditions and poses. The results are summarized at Table 1, where it is presented that TNN-MO model exhibits the position error of 0.204 m and the attitude error of 0.91 deg. The position error is about 0.8% of the maximum range. The level of error exhibited by the proposed TNN-MO is comparable to that of common visual-inertial odometry techniques [28], although a direct comparison is not feasible due to differing assumptions. When compared with the TNN-SO model, the proposed TNN-MO model demonstrates improved accuracy, even over long ranges. From Fig. 14, we can see that the TNN-SO model is unable to detect the keypoints of the dog house when the camera is far from the ship. However, our TNN-MO model is capable of detecting keypoints of multiple objects and it is not vulnerable to the rot., rotation; pos., position; est., Estimation.

**Table 1 Validation with synthetic data set**

| TNN type | No. of images | Max range, $L$, m | MAE rot. est., deg | MAE pos. est., m | MAE/$L$, % |
|---|---|---|---|---|---|
| TNN-SO | 4,200 | 20 | 1.13 | 0.281 | 1.41 |
| TNN-MO | 5,500 | 25 | 0.91 | 0.204 | 0.82 |

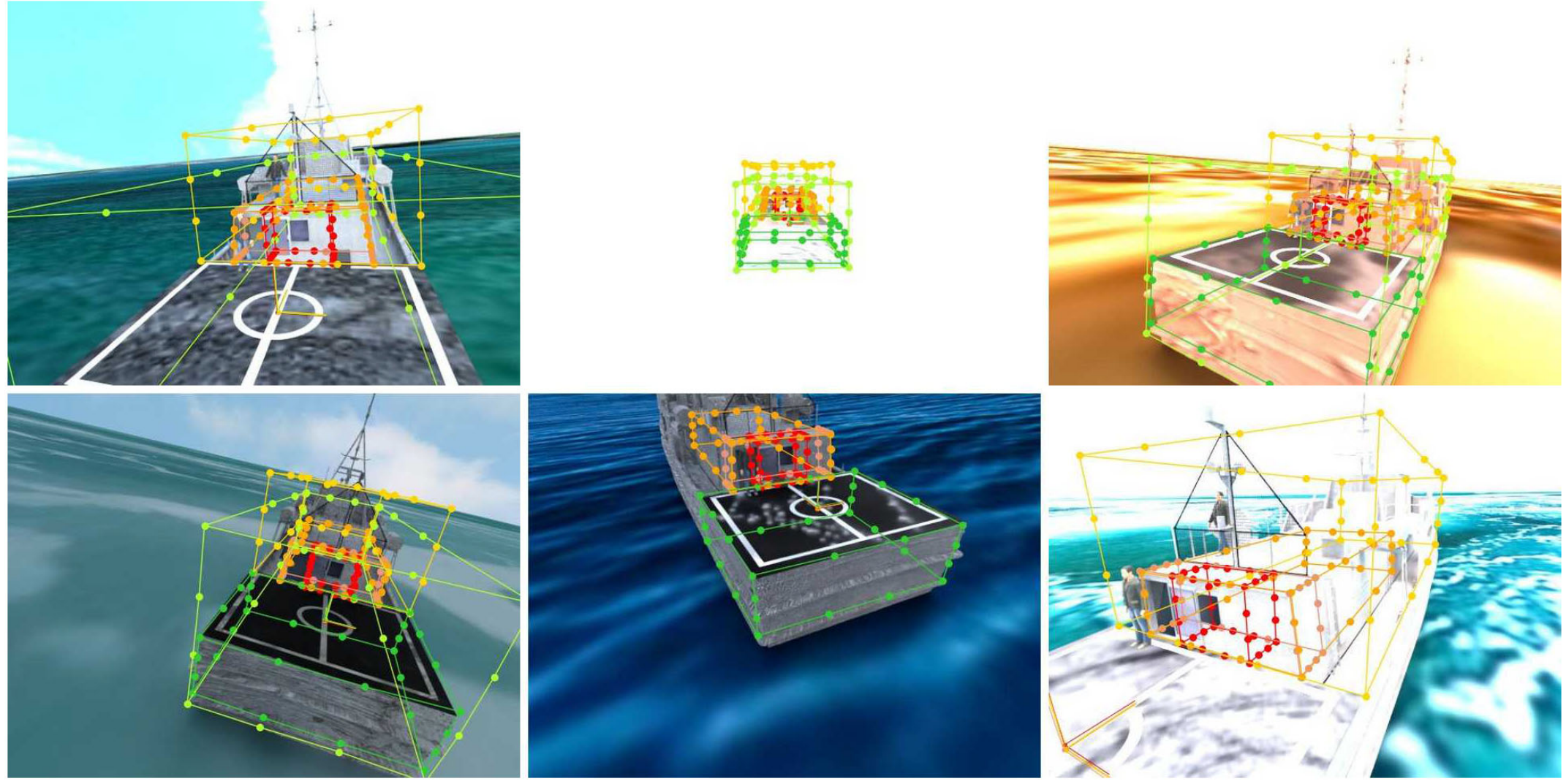

**Fig. 13 TNN-MO model tested over synthetic data: for object classes with the confidence $o_j > 0.9$, the key points and the base frame are illustrated in colors.**

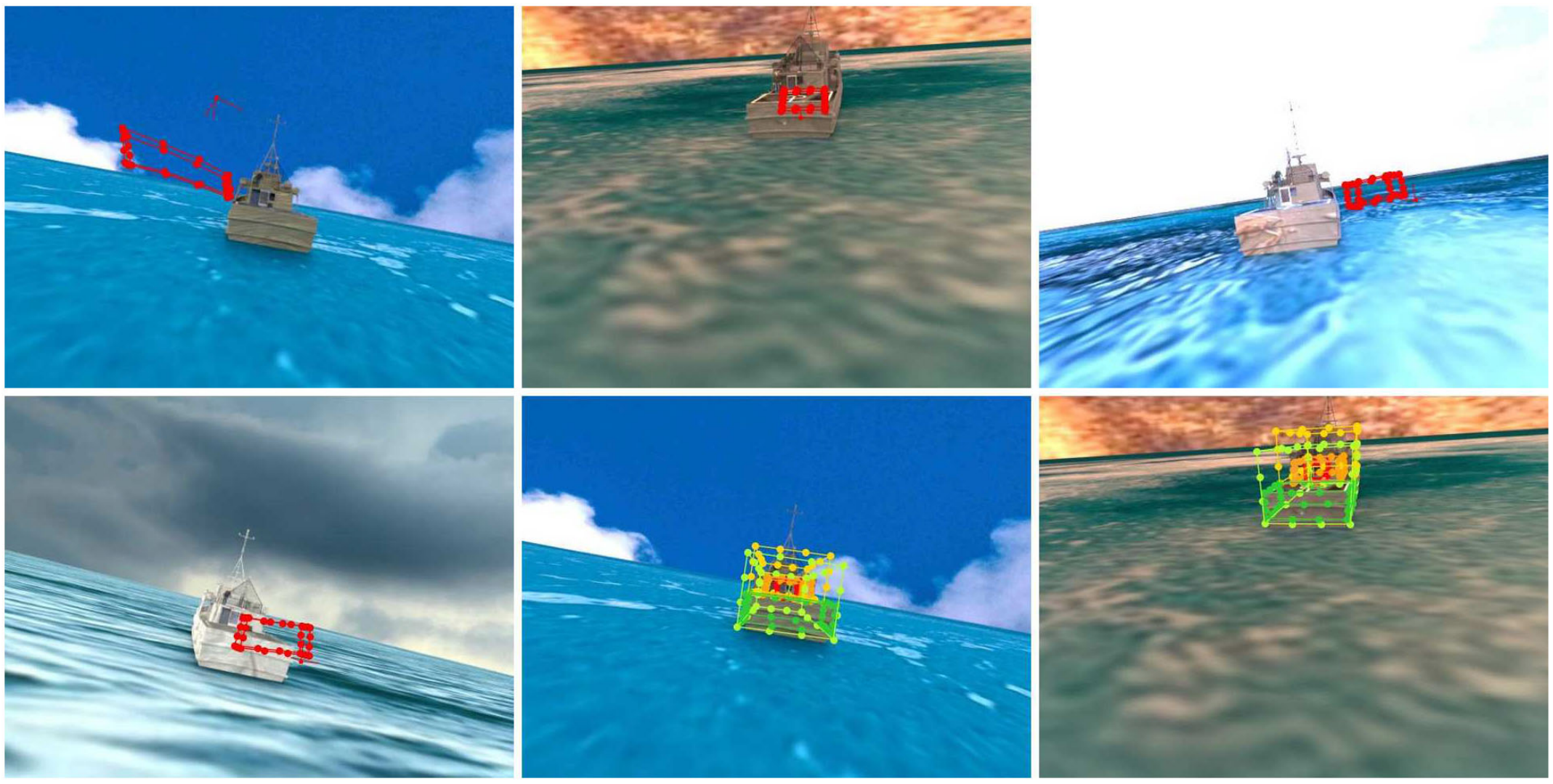

**Fig. 14 TNN-SO (top) vs TNN-MO (bottom) tested over long range synthetic data: for TNN-MO model, object classes with the confidence $o_j > 0.9$, the key points are illustrated in colors.**

visibility of a single object. Figures 13 and 14 illustrate the bounding boxes with a higher confidence level computed from the estimated pose, and they demonstrate that the estimated key-points from the TNN-MO model exhibits the grater accuracy over a longer range.

## VI. Validation with Flight Experiments

The TNN-MO model, trained on a synthetic data set, was further tested using real images obtained during flight experiments on a USNA research vessel. The model was trained on a synthetic data set, as collecting a large labeled data set of a research vessel in the ocean is infeasible. As such, it is critical to evaluate the model in the real-world situation to access its capacity to handle the unique attributes of real images, such as such as variations in lighting conditions, object appearances, backgrounds, and other factors that may deviate from the synthetic data set, referred to as the sim-to-real gap. This will offer important insights into its adaptability and resilience, confirming its efficacy beyond the synthetic training data and its suitability for real-world applications.

### A. Data Collection System

The hardware configuration to capture a validation data set is as follows. A data collection system (DCS) as shown in Fig. 16, that was developed for ship air wake measurements [15], is augmented with a camera. It is attached to an octocoper unmanned aerial vehicle that is manually controlled to fly around a USNA research vessel in Chesapeake Bay, Maryland.

The detailed configuration of the DCS composed of a base module and a rover module is presented at Fig. 15. Specifically, the DCS includes inertial measurement units (IMUs) and RTK GPS, from which an extended Kalman filter is executed to estimate the pose in real time. The rover module of the DCS has been enhanced with an Alvium 1800C-240 Global Shutter RGB camera and a Jetson Nano single-board computer, connected via a MIPI (Mobile Industry Processor Interface) CSI-2 (Camera Serial Interface 2) cable. This setup allows image capture at a rate of 5 Hz, synchronized with the RTK GPS via a Wi-Fi connection. This upgraded rover configuration enables synchronization and time stamping of the images within the RTK GPS data collection process. It allows for the effective collection of real image data along with accurate relative camera 6D poses.

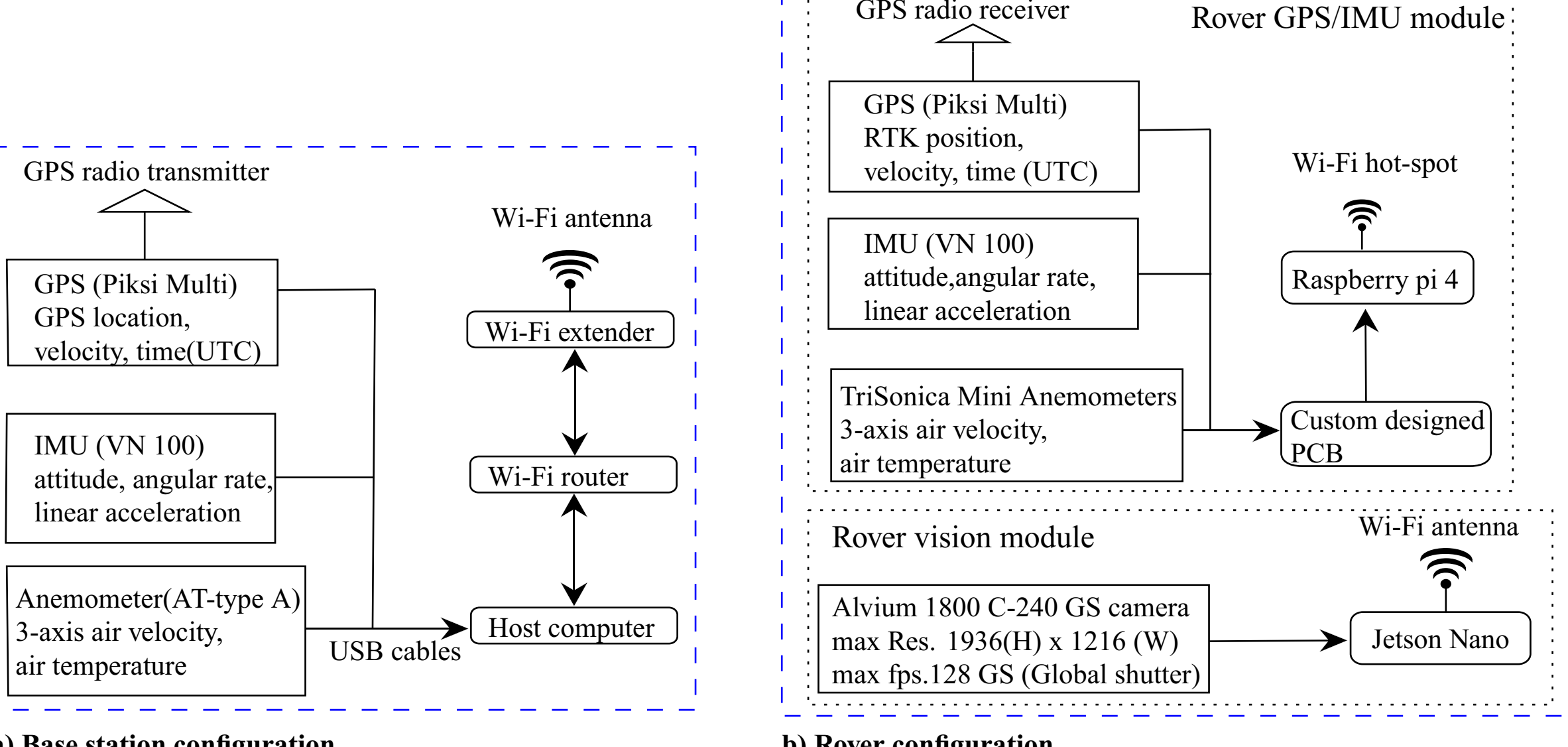

**Fig. 15 Data Collection System. VN: VectorNav; UTC: Coordinated Universal Time; Res.: Resolution; fps.: Frames per second; PCB: Printed Circuit Board. (AT-type A refers to a specific model name or manufacturer designation).**

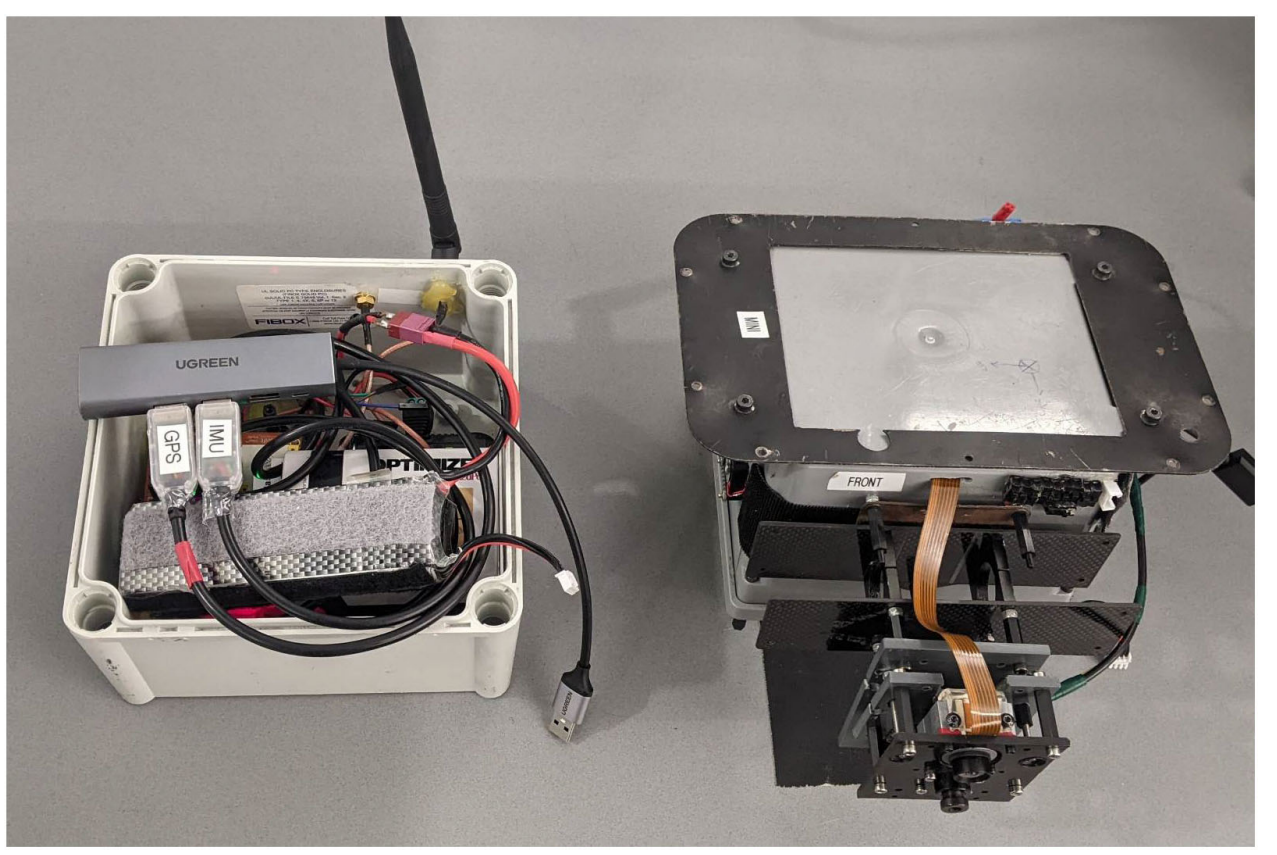

**Fig. 16 Base (left) and camera fixed rover (right) setup used of the DCS.**

### B. Model Testing with Real Data

The trained model is tested with the real-world images captured by the DCS over multiple days. During the in-flight experiments conducted on the USNA research vessel, the model was subjected to real-world conditions where instances of variable lighting and occlusions were encountered. We deliberately selected challenging images for testing because they represent realistic conditions where such situations can occur. Figures 17–19 show the keypoints prediction and corresponding re-projected ships coordinate frame from estimated pose by the proposed TNN-MO model under the following three lighting conditions:

1) An *overexposed ship* is one where the image has captured too much light, causing the ship to appear excessively bright. This results in a loss of detail, especially in areas that have subtle color variations or textures. The ship's features become hard to be distinguished because the intense light overwhelms the camera's sensor, leading to a predominance of white or light areas, particularly on surfaces like the landing pad. It is as if the ship is caught in a glare, with its details bleached by the brightness.

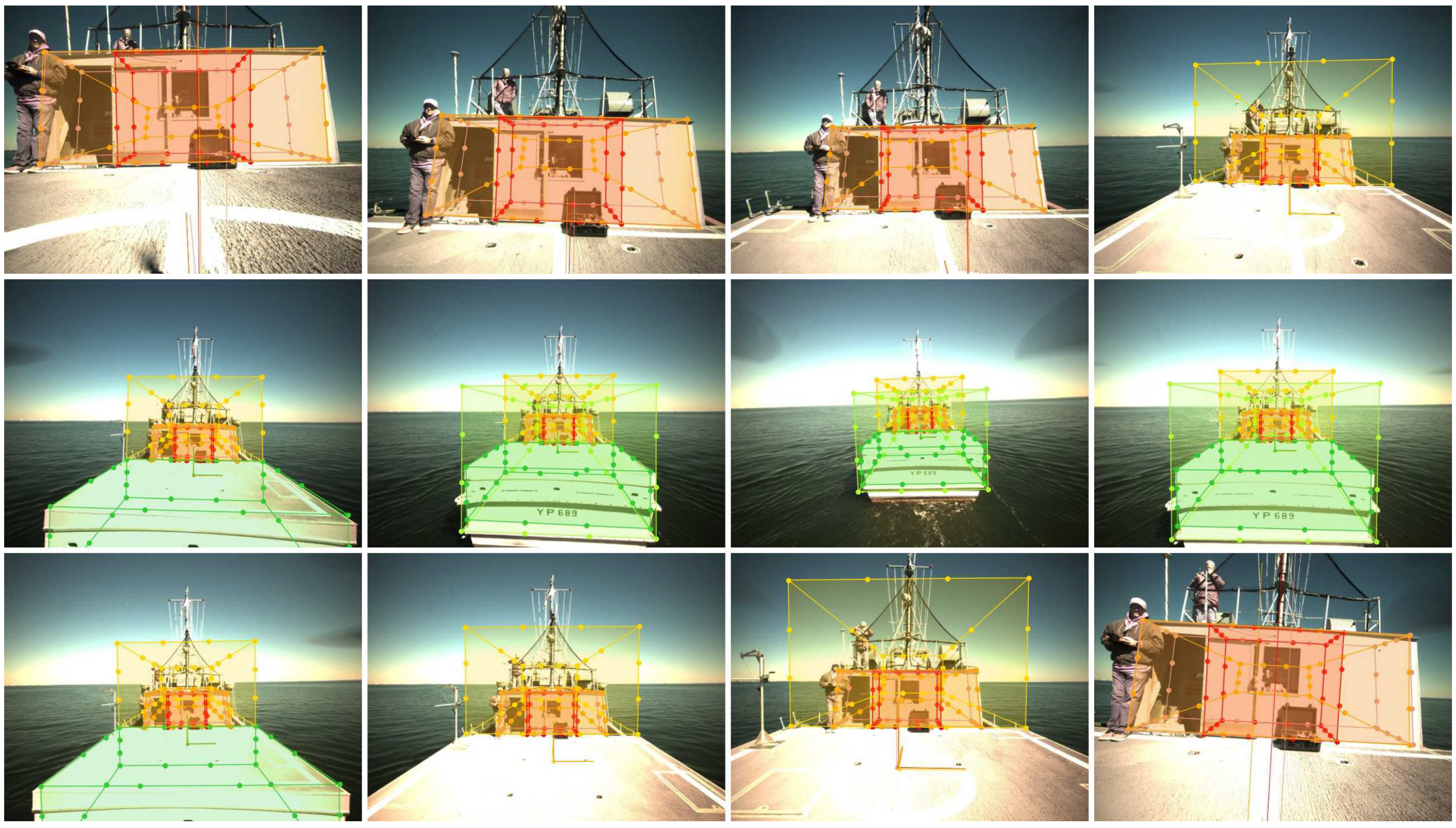

**Fig. 17 Estimated keypoints and bounding boxes for overexposed images: objects with the confidence $o_j > 0.9$ are highlighted.**

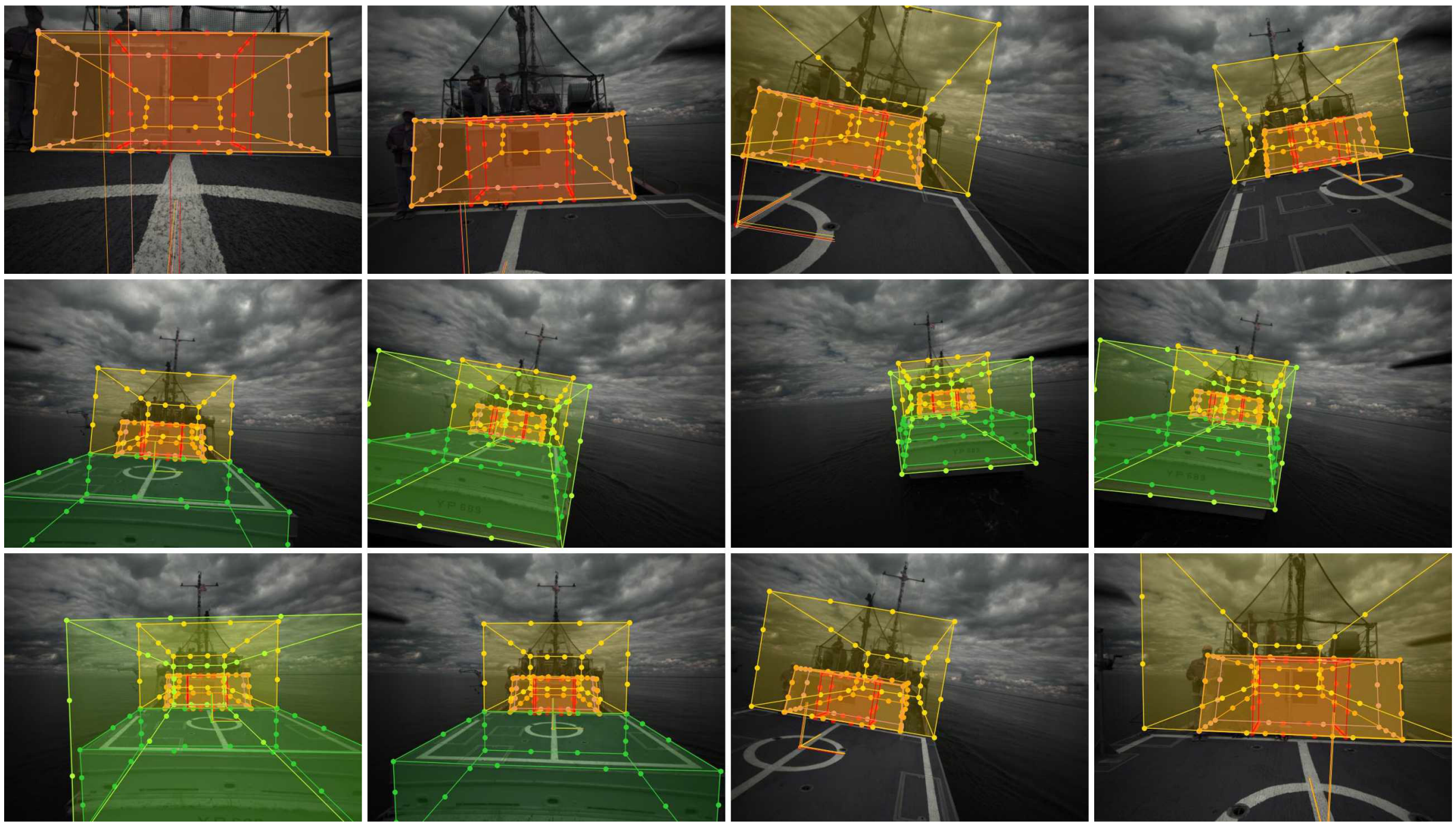

**Fig. 18 Estimated keypoints and bounding boxes for underexposed images: objects with the confidence $o_j > 0.9$ are highlighted.**

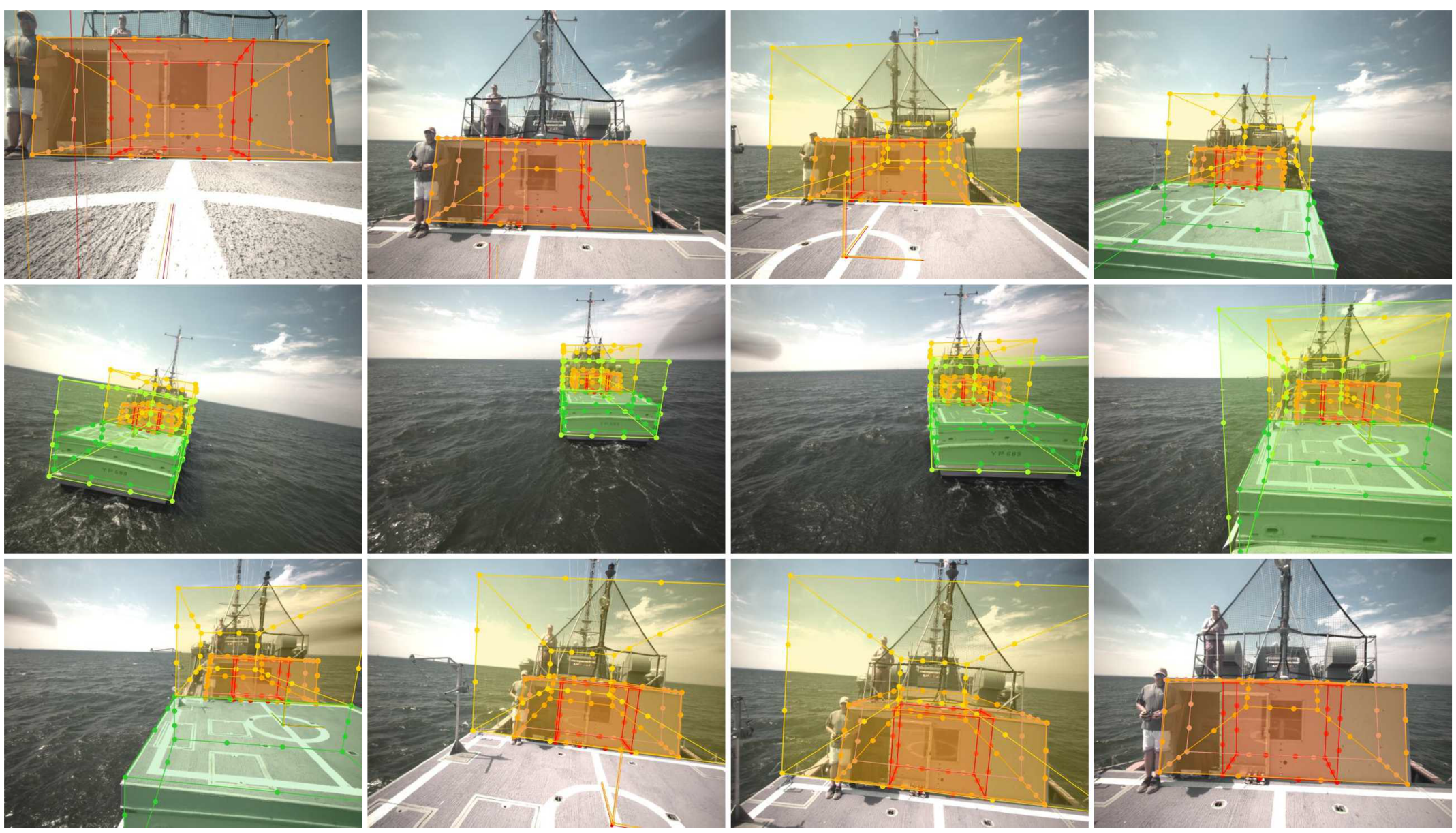

**Fig. 19 Estimated keypoints and bounding boxes for normal images: objects with the confidence $o_j > 0.9$ are highlighted.**

2) An *underexposed ship* is one where the image has not captured enough light, making the ship appear too dark. This can obscure details and make it challenging to distinguish features, especially in areas that are naturally shadowed or lack reflective surfaces. It is as if the ship is enveloped in shadows, with its details concealed in darkness.

3) A *normal ship*, in terms of exposure, is one where the lighting conditions are optimal for image analysis, resulting in a balanced image with clear visibility of details. The lighting is neither too intense nor too dim, providing an optimal level of brightness that allows all parts of the ship to be clearly distinguished.

Traditional feature extraction methods and visual marker-based approaches often struggle under challenging circumstances, as there is a lack of visual features on the ship.

Furthermore, in this dynamic environment of real-world imaging, human operators and various items are present on the ship, which we can identify as occlusions. These occlusions, along with variable lighting, present significant challenges. However, as illustrated by Figs. 17–19, the proposed model successfully identified the keypoints and the bounding boxes of multiple parts of the ship. Its robust performance under such conditions is a testament to its capability to provide accurate results, ensuring reliable 6D pose estimation even when faced with obstructions and fluctuating light levels.

### C. Validation with RTK GPS

Further, for quantitative validation, the estimated pose is compared with the actual pose determined by the DCS. The DCS integrates the relative attitude, derived from the base rover's IMUs, and the relative position from the RTK-GPS with an extended Kalman filter. The IMUs, such as the VN-100, provide a magnetic heading accuracy of 2.0°, a gyro in-run bias stability of 5°/h, and a pitch/roll accuracy of 0.5° under normal conditions. The RTK-GPS, such as the Piksi Multi Global Navigation Satellite System (GNSS), provides a horizontal position accuracy of 0.75 m in Satellite-Based Augmentation System (SBAS) mode, a velocity accuracy of 0.03 m/s rms, and a time accuracy of 60 ns rms. In the RTK mode, it provides an accuracy of 0.010 m horizontally and 0.015 m vertically. These systems provide a centimeter-level accuracy in nominal conditions for the DCS. Here, the measurements of the RTK-GPS are considered as ground truth. The trajectories estimated by the proposed TNN-MO model and the RTK-GPS under the three mentioned illumination conditions are illustrated in Figs. 20–22 with the estimation errors and $3\sigma$ bounds calculated by Eqs. (4) and (9), where it is shown that the position and the attitude trajectory estimated by the TNN-MO model are consistent with the IMU and the RTK-GPS.

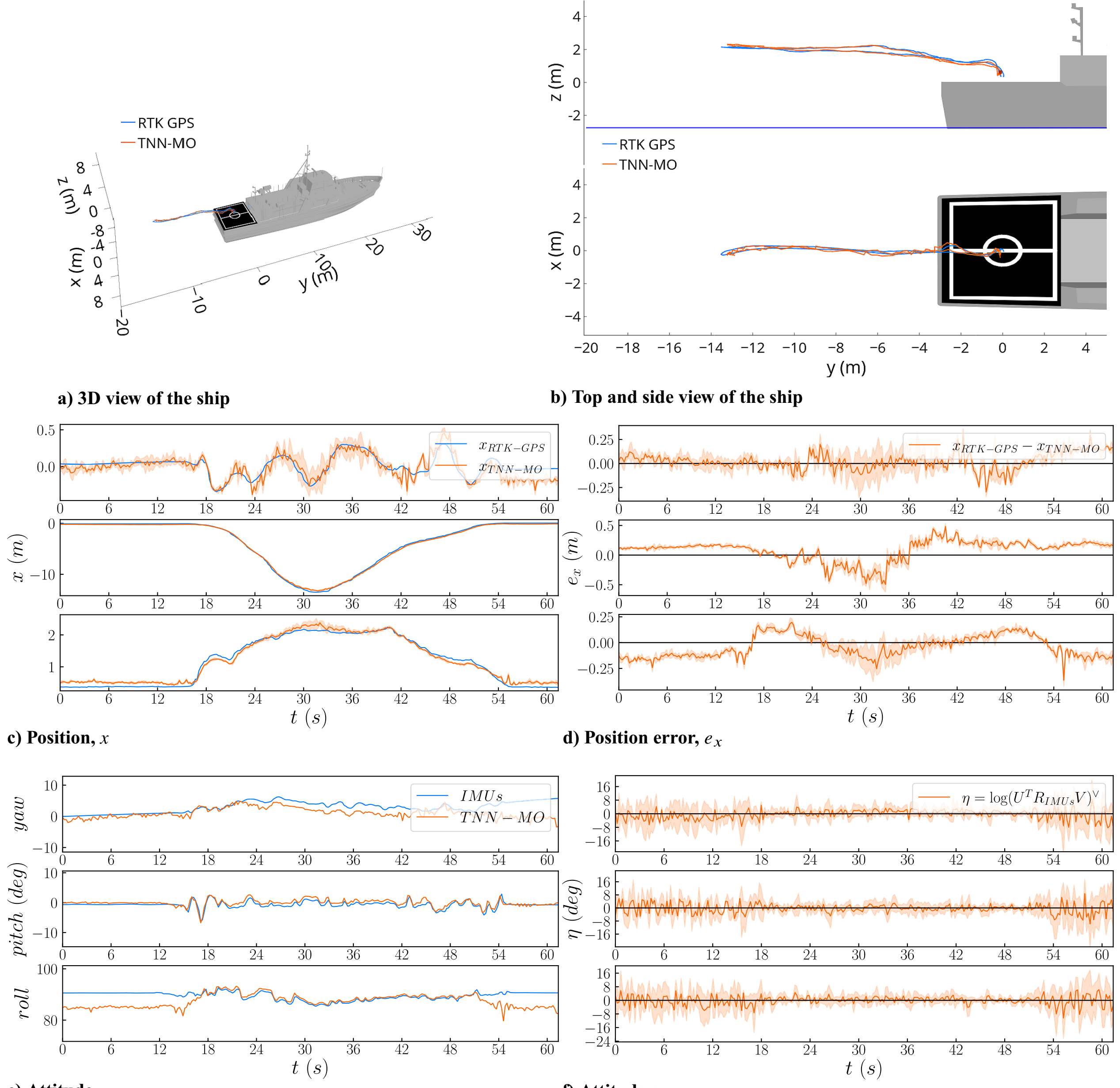

a) 3D view of the ship

b) Top and side view of the ship

c) Position, $x$

d) Position error, $e_x$

e) Attitude, $ypr$

f) Attitude error, $\eta$

**Fig. 20 Overexposed case: the estimated pose (red) is compared against the RTK-GPS measurements (blue).**

The mean absolute errors are also tabulated at Table 2. The MAE of position estimation ranges from 0.089 to 0.177 m, which corresponds to 0.66–0.97% of the maximum range $L$. The MAE of rotation estimation varies from 1.1 to 4.0 deg. Although these errors are slightly larger than the validation error for the synthetic data as presented in Table 1, it is verified that the proposed TNN-MO successfully overcome the sim-to-real gap. The performance is consistent over the varying illumination conditions captured over multiple days. Although the degree of attitude uncertainties measured by the standard deviation is in the reasonable range, the standard deviation for the position estimate is relatively small compared with the error, indicating overconfidence. This is partially because the uncertainties are measured indirectly in Eq. (4) by the variations of the estimate over multiple objects, not in the confidence of estimate for the individual object. However, in Figs. 23–25, the uncertainties increase as the camera is farther away from the ship, which is expected. These show that the presented uncertainty model behaves reasonably in the qualitative sense, but not necessarily accurate quantitatively. The exact modeling of uncertainties is considered as one of future directions. This provides promising results for pose estimate that can be used for autonomous launch and recovery in the ocean environments.

### D. Multi-Objects Pose Estimation

The proposed TNN-MO model integrates the multiple pose estimates with respect to six object classes. To investigate the advantages of the multi-objects estimates, we perform an ablation study by studying the pose estimated with respect to the individual object class without fusion.

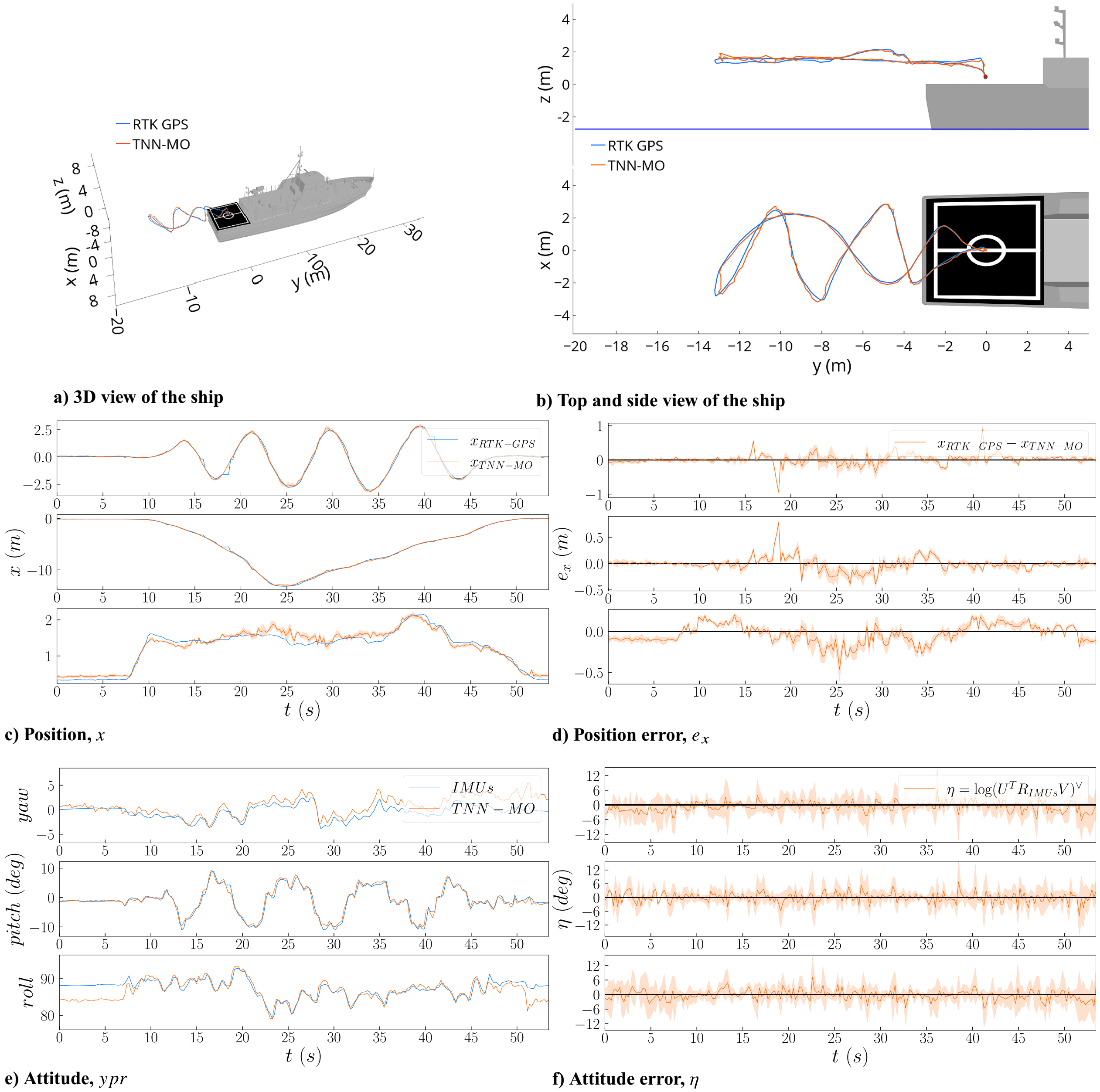

a) 3D view of the ship

b) Top and side view of the ship

c) Position, $x$

d) Position error, $e_x$

e) Attitude, $ypr$

f) Attitude error, $\eta$

**Fig. 21 Underexposed case: the estimated pose (red) is compared against the RTK-GPS measurements (blue).**

For the second case of the underexposed ship, the trajectory estimated by each object is presented at Fig. 23. Upon examining Figs. 23d–23f, it is evident that when the camera is closer to the ship, three object categories of the superstructure, the whole ship, and the stern are not within the field of view, and the corresponding estimate is degraded. Conversely, for Figs. 23a–23c, object pose estimations are accurate and reliable, as the objects remain visible throughout the flight.

Furthermore, the position estimation error with respect to each object is summarized at Table 3, in which it is shown that the proposed multi-object approach yields smaller errors in mean and max values compared to the TNN-SO, illustrating the advantages of fusion.

### E. Attention Map

To gain insights into how the synthetically trained TNN-MO model performs on real-world images, we analyzed the visualizations of the attention maps for the encoder and the decoder. The self-attention mechanism in the transformer network enables the encoder to understand the context of image elements. This is crucial in object detection, where the relationship between different parts of the image can provide important information. Therefore, we analyzed the transformer self-attention maps by passing a real-world image into TNN-MO. In Fig. 24, the encoder's self-attention for a selected set of reference points is visualized in the attention map using 300 ($H_0 \times W_0$) features from the convoluted input image. Each pixel in the image corresponds to one of these features. We are visualizing how a feature linked to a selected pixel attends to other features. By analyzing the self-attention map, we can see that the encoder is capable of identifying key elements such as the sea, sky, and ship within the image based on the selected points. We observed that the TNN-MO encoder's self-attention mechanism allows the model to understand the context of each pixel in relation to the others. For example, the self-attention map for the pixel point on the ship in the image, attention weights gets higher in the similar area of the ship on the map.

Additionally, to gain insights into what the model is focusing on, we visualized the decoder cross-attention map for different queries. In Fig. 25, we are visualizing how a encoder output

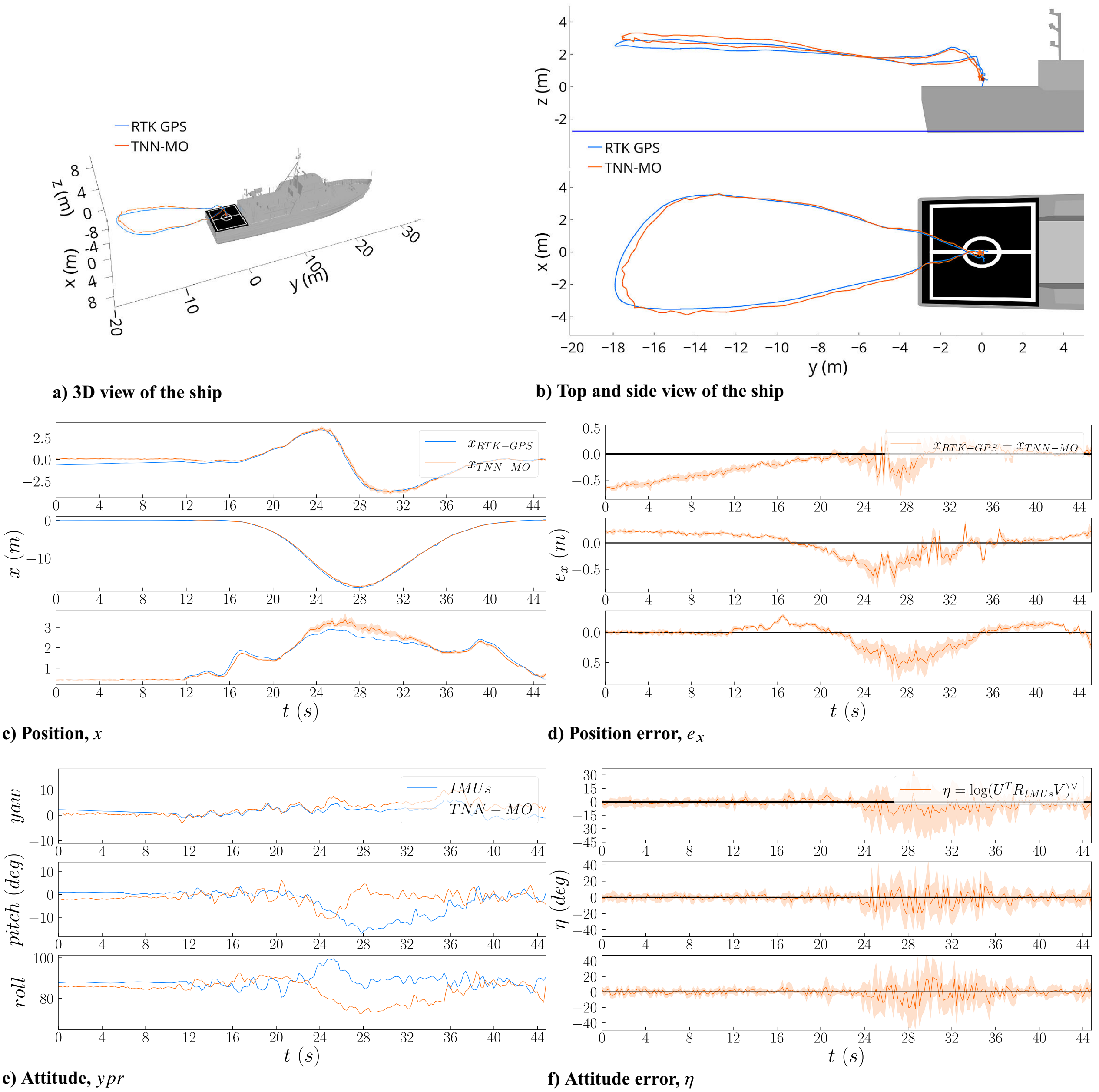

a) 3D view of the ship

b) Top and side view of the ship

c) Position, $x$

d) Position error, $e_x$

e) Attitude, $ypr$

f) Attitude error, $\eta$

**Fig. 22 Nominal case: the estimated pose (red) is compared against the RTK-GPS measurements (blue).**

**Table 2 Accuracy of 6D pose estimation for TNN-MO model under variable lighting conditions**

| Image type | Max range, $L$, m | MAE/$\sigma$/$d$ of rot., deg | MAE/$\sigma$ of pos., m | MAE/$L$, % |
|---|---|---|---|---|
| Overexposed ship | 13.7 | 1.8/2.32/0.999 | 0.112/0.017 | 0.82 |
| Underexposed ship | 13.5 | 1.1/1.83/0.999 | 0.089/0.019 | 0.66 |
| Normal ship | 18.2 | 4.0/4.54/0.999 | 0.177/0.022 | 0.97 |

rot., rotation; pos., position.

attends to each queries related to the objects defined in Fig. 9, excluding the cross-attention related to seventh query, which is for the ∅ object class. This can help us understand which parts of the image the model finds most relevant for each query, providing valuable insights into the model's decision-making process. We observed that the area where the cross-attention weight is higher for each query lies in the ship area of the image. This means that our TNN-MO model is able to understand the relevant information coming from the encoder to identify each part of the ship.

## VII. Conclusions

This study introduces an innovative method for estimating the relative 6D pose of an autonomous UAV in relation to a ship, using monocular RGB camera images. The authors developed a synthetic data set incorporating varied textures, lighting conditions, and camera poses and annotated them with 2D keypoints of ship components in a virtual environment. Subsequently, we present a network architecture based on the transformer network, trained to detect keypoints of multiple parts of the ship, from which the camera pose is estimated using the EPnP algorithm. Then, the estimated 6D poses are filtered, where the object class confidence is greater than 0.9, and they are integrated with Bayesian fusion to provide reliable pose estimations. This process combines pose estimations from multiple objects, disregarding information from unreliable estimates and placing more reliance on credible ones. The trained model, thoroughly tested and validated on both synthetic and real-world data, yielded mean absolute position errors of 0.8% and 1.0% of the maximum flight range, respectively.

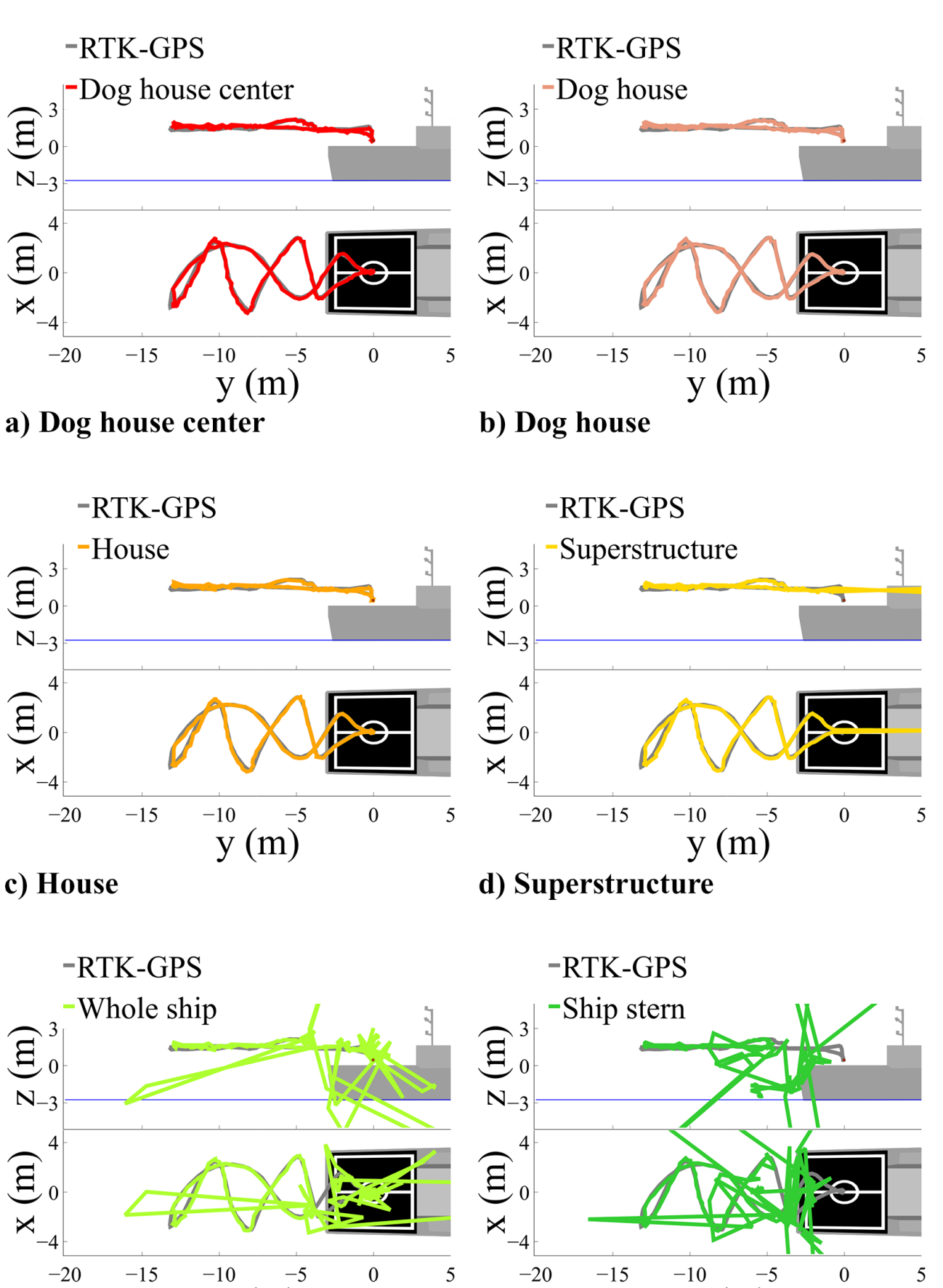

a) Dog house center b) Dog house c) House d) Superstructure e) Whole ship f) Ship stern

**Fig. 23 Pose estimated with respect to the individual object class without fusion for underexposed images.**

This proposed formulation, based on synthetic data, circumvents challenges associated with collecting a large set of labeled data in the real world, successfully overcoming the discrepancy between virtual and real environments. Furthermore, the proposed multi-object pose estimation and fusion enhance accuracy and robustness under various relative configurations, occlusions, and illumination conditions. These results underscore the suitability of our approach for improving the autonomy and safety of UAV operations, particularly for autonomous landing and takeoff on moving platforms.

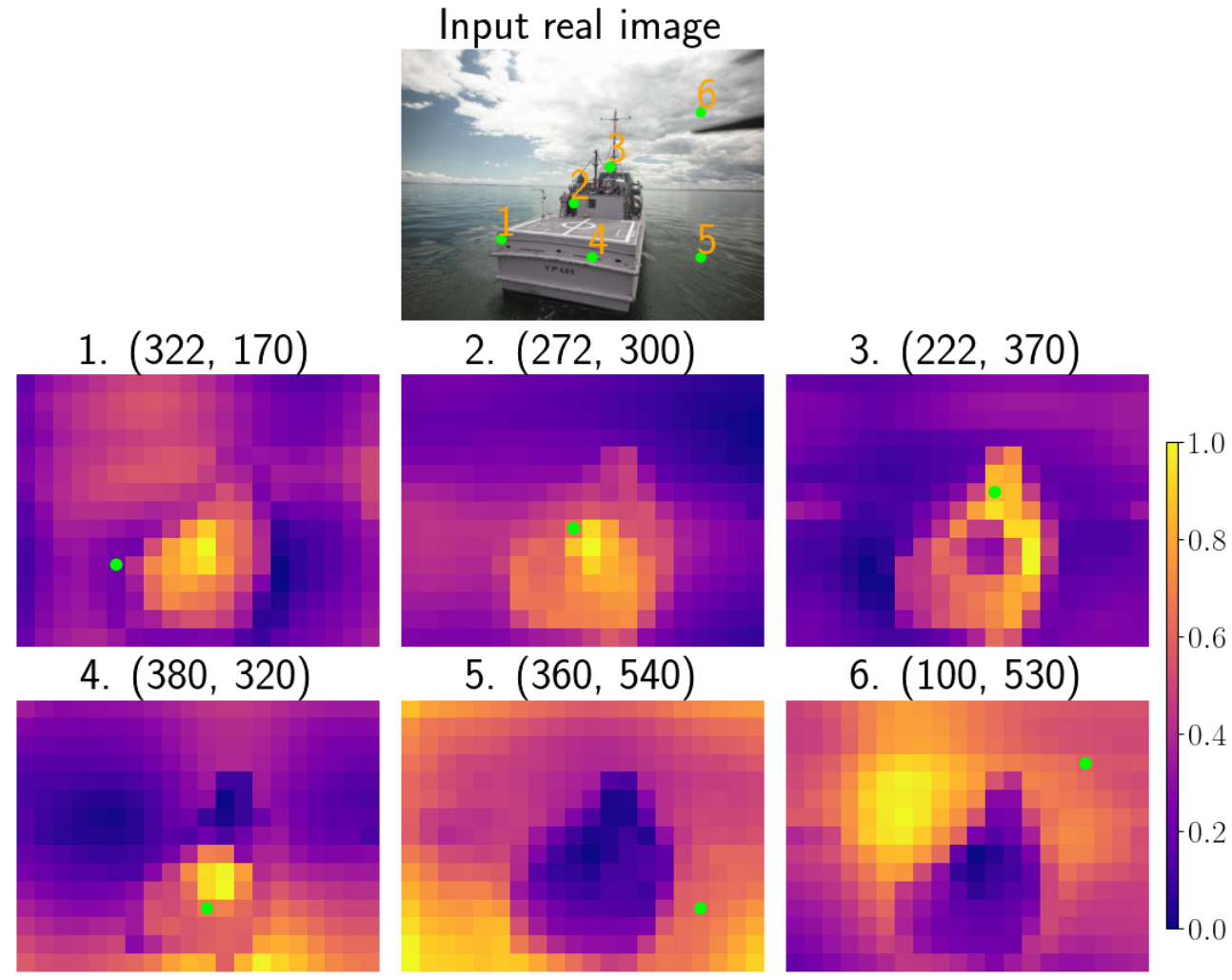

**Fig. 24 Encoder self-attention for a selected set of reference points: the attention map by using 300 ($H_0 \times W_0$) features from the convoluted input image.**

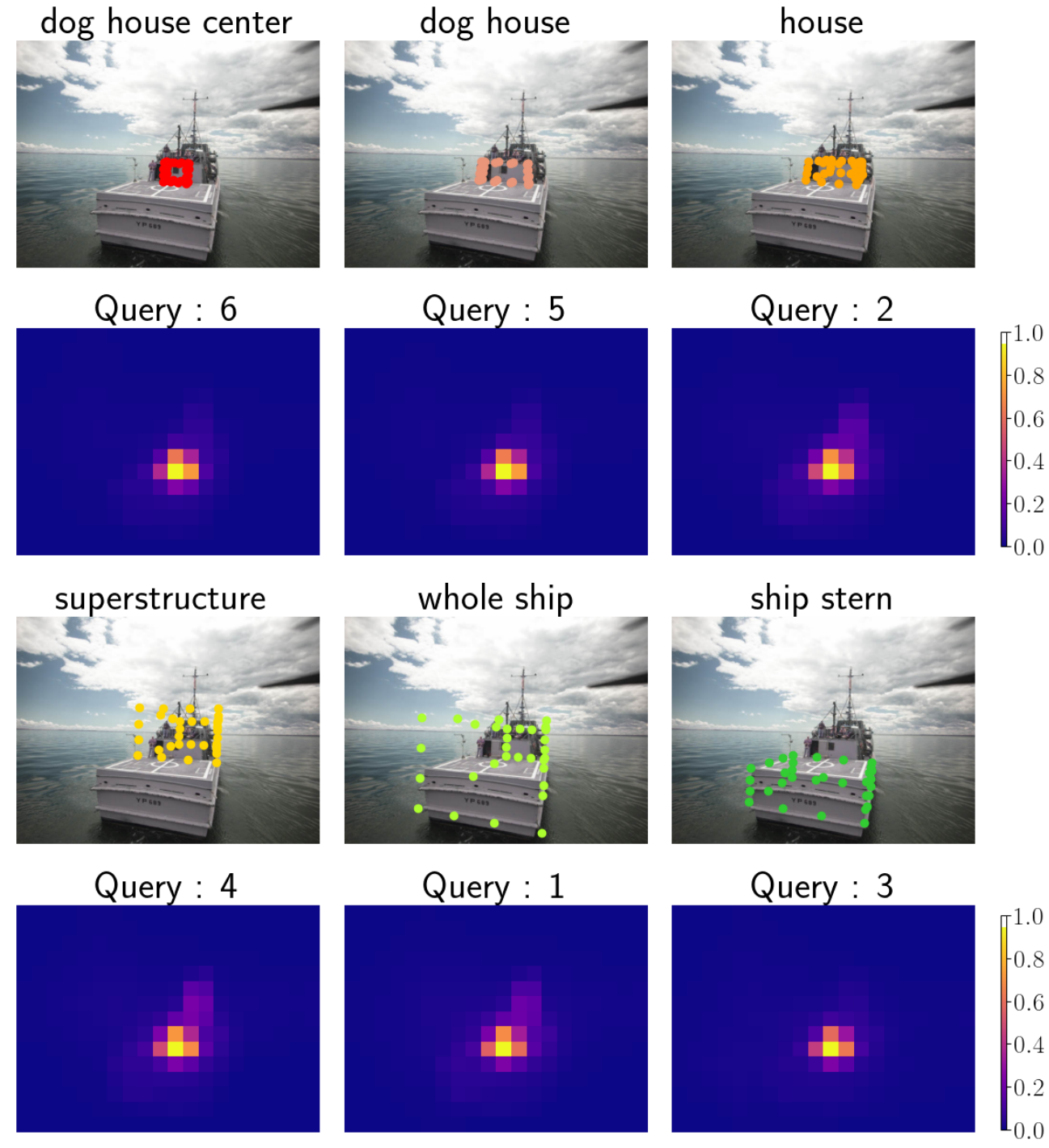

**Fig. 25 Object keypoints are predicted by TNN-MO in the given real image (top), and the decoder cross-attention maps for the object queries corresponding to the predictions in the first row are shown (bottom).**

**Table 3 Position estimation error for TNN-MO and TNN-SO**

| | Overexposed ship | | Underexposed ship | | Normal ship | |
|---|---|---|---|---|---|---|
| | Pos. err., m | | Pos. err., m | | Pos. err., m | |
| | Mean | Max | Mean | Max | Mean | Max |
| TNN-MO | **0.112** | **0.485** | **0.089** | **0.937** | **0.177** | **0.666** |
| TNN-SO (dog house center) | 0.129 | 0.723 | 0.259 | 1.669 | 0.261 | 1.191 |
| TNN-SO (dog house) | 0.131 | 0.720 | 0.259 | 1.738 | 0.253 | 1.154 |
| TNN-SO (house) | 0.123 | 0.728 | 0.258 | 1.667 | 0.254 | 1.220 |
| TNN-SO (superstructure) | 0.135 | 0.782 | 0.347 | 1.660 | 0.304 | 1.169 |
| TNN-SO (whole ship) | 0.151 | 0.682 | 0.369 | 1.704 | 0.355 | 1.129 |
| TNN-SO (ship stern) | 0.152 | 0.669 | 0.394 | 1.664 | 0.365 | 1.130 |

err., error; pos., position.

Future work includes integrating the pose estimate of the proposed transformer network model with an inertial measurement unit for visual-inertial navigation and using it for autonomous flight experiments in ocean environments.

## Acknowledgments

This research was conducted on the U.S. Naval Academy's research vessel YP689, and we extend our gratitude to the U.S. Naval Academy and the crew of YP689. The research has been partially supported by the U.S. Naval Academy / Naval Supply Systems Command (USNA/NAVSUP, Grant No. N0016123RC01EA5), the National Science Foundation (NSF, Grant No. CNS-1837382), the Air Force Office of Scientific Research Multidisciplinary University Research Initiative (AFOSR MURI, Grant No. FA9550-23-1-0400), and the Office of Naval Research (ONR, Grant No. N00014-23-1-2850). The authors express their sincere gratitude for this support.

## References

[1] Ngo, T. D., and Sultan, C., "Model Predictive Control for Helicopter Shipboard Operations in the Ship Airwakes," *Journal of Guidance, Control, and Dynamics*, Vol. 39, No. 3, 2016, pp. 574–589. https://doi.org/10.2514/1.G001243

[2] Abu-Jbara, K., Sundaramorthi, G., and Claudel, C., "Fusing Vision and Inertial Sensors for Robust Runway Detection and Tracking," *Journal of Guidance, Control, and Dynamics*, Vol. 41, No. 9, 2018, pp. 1929–1946. https://doi.org/10.2514/1.G002898

[3] Dougherty, J. A., and Lee, T., "Monocular Estimation of Ground Orientation for Autonomous Landing of a Quadrotor," *Journal of Guidance, Control, and Dynamics*, Vol. 39, No. 6, 2016, pp. 1407–1416. https://doi.org/10.2514/1.G001229

[4] Chatterji, G. B., Menon, P. K., and Sridhar, B., "Vision-Based Position and Attitude Determination for Aircraft Night Landing," *Journal of Guidance, Control, and Dynamics*, Vol. 21, No. 1, 1998, pp. 84–92. https://doi.org/10.2514/2.4201

[5] Marut, A., Wojtowicz, K., and Falkowski, K., "ArUco Markers Pose Estimation in UAV Landing Aid System," *2019 IEEE 5th International Workshop on Metrology for AeroSpace (MetroAeroSpace)*, Inst. of Electrical and Electronics Engineers, New York, 2019, pp. 261–266. https://doi.org/10.1109/MetroAeroSpace.2019.8869572

[6] Lu, Q., Ren, B., and Parameswaran, S., "Shipboard Landing Control Enabled by an Uncertainty and Disturbance Estimator," *Journal of Guidance, Control, and Dynamics*, Vol. 41, No. 7, 2018, pp. 1502–1520. https://doi.org/10.2514/1.G003073

[7] Robaglia, A. E., Libine, S., and Gamagedara, K., "Autonomous Landing of an Unmanned Aerial Vehicle on a Moving Ship," AIAA Paper 2018-1461, 2018. https://doi.org/10.2514/6.2018-1461

[8] Gamagedara, K., Lee, T., and Snyder, M., "Delayed Kalman Filter for Vision-Based Autonomous Flight in Ocean Environments," *Control Engineering Practice*, Vol. 143, 2024, Paper 105791. https://doi.org/10.1016/j.conengprac.2023.105791

[9] Kang, J., Liu, W., Tu, W., and Yang, L., "YOLO-6D+: Single Shot 6D Pose Estimation Using Privileged Silhouette Information," *2020 International Conference on Image Processing and Robotics (ICIP)*, IEEE Publ., Piscataway, NJ, 2020, pp. 1–6. https://doi.org/10.1109/ICIP48927.2020.9367354

[10] Wang, G., Manhardt, F., Tombari, F., and Ji, X., "GDR-Net: Geometry-Guided Direct Regression Network for Monocular 6D Object Pose Estimation," *CoRR*, Vol. abs/2102.12145, 2021, pp. 16,606–16,616. https://doi.org/10.1109/CVPR46437.2021.01634

[11] Carion, N., Massa, F., Synnaeve, G., Usunier, N., Kirillov, A., and Zagoruyko, S., "End-to-End Object Detection with Transformers," *European Conference on Computer Vision (ECCV)*, Vol. 12346, Lecture Notes in Computer Science, Springer, Cham, 2020, pp. 213–229. https://doi.org/10.1007/978-3-030-58452-8_13

[12] Amini, A., Periyasamy, A. S., and Behnke, S., "YOLOPose: Transformer-Based Multi-Object 6D Pose Estimation Using Keypoint Regression," *Intelligent Autonomous Systems 17*, edited by I. Petrovic, E. Menegatti, and I. Marković, Springer Nature Switzerland, Cham, 2023, pp. 392–406. https://doi.org/10.1007/978-3-031-22216-0_27

[13] Wickramasuriya, M., Lee, T., and Snyder, M., "Deep Monocular Relative 6D Pose Estimation for Ship-Based Autonomous UAV," *AIAA SCITECH 2024 Forum*, AIAA Paper 2024-2877, 2024. https://doi.org/10.2514/6.2024-2877

[14] Lepetit, V., Moreno-Noguer, F., and Fua, P., "EPnP: An Accurate O(n) Solution to the PnP Problem," *International Journal of Computer Vision*, Vol. 81, No. 2, 2009, pp. 155–166. https://doi.org/10.1007/s11263-008-0152-6

[15] Gamagedara, K., Lee, T., and Snyder, M. R., "Real-Time Kinematics GPS Based Telemetry System for Airborne Measurements of Ship Air Wake," AIAA Paper 2019-2377, 2019. https://doi.org/10.2514/6.2019-2377

[16] Bostock, N., Richez, A., Costello, D. H., Webster-Giddings, A., and Wickramasuriya, M., "Verification of YP689 Flow Field Models for Dynamic Interface Flight Test," *AIAA SCITECH 2023 Forum*, AIAA Paper 2023-0294, 2023. https://doi.org/10.2514/6.2023-0294

[17] Schönberger, J. L., and Frahm, J.-M., "Structure-from-Motion Revisited," *Conference on Computer Vision and Pattern Recognition (CVPR)*, Inst. of Electrical and Electronics Engineers, New York, 2016, pp. 4104–4113. https://doi.org/10.1109/CVPR.2016.445

[18] Blender Online Community, *Blender—A 3D Modelling and Rendering Package*, Ver. 3.3, Blender Foundation, Amsterdam, 2022, http://www.blender.org.

[19] Loquercio, A., Kaufmann, E., Ranftl, R., Dosovitskiy, A., Koltun, V., and Scaramuzza, D., "Deep Drone Racing: From Simulation to Reality with Domain Randomization," *IEEE Transactions on Robotics*, Vol. 36, No. 1, 2019, pp. 1–14. https://doi.org/10.1109/TRO.2019.2942989

[20] Cimpoi, M., Maji, S., Kokkinos, I., Mohamed, S., and Vedaldi, A., "Describing Textures in the Wild," *Proceedings of the IEEE Conference on Computer Vision and Pattern Recognition (CVPR)*, IEEE Publ., Piscataway, NJ, 2014, pp. 3606–3613.

[21] "OpenGameArt," n.d, https://opengameart.org/ [retrieved 12 May 2023].

[22] Budinich, R., "A Region Based Easy Path Wavelet Transform for Sparse Image Representation," *International Journal of Wavelets, Multiresolution and Information Processing*, Vol. 15, No. 05, 2017, Paper 1750045. https://doi.org/10.1142/S021969131750045X

[23] Vaswani, A., Shazeer, N., Parmar, N., Uszkoreit, J., Jones, L., Gomez, A. N., Kaiser, L., and Polosukhin, I., "Attention Is All You Need," *Advances in Neural Information Processing Systems*, edited by I. Guyon, U. V. Luxburg, S. Bengio, H. Wallach, and R. Fergus, Vol. 30, Curran Assoc., New York, 2017. https://doi.org/10.48550/arXiv.1706.03762

[24] Stewart, R., Andriluka, M., and Ng, A. Y., "End-to-End People Detection in Crowded Scenes," *2016 IEEE Conference on Computer Vision and Pattern Recognition (CVPR)*, Inst. of Electrical and Electronics Engineers, New York, 2016, pp. 2325–2333. https://doi.org/10.1109/CVPR.2016.255

[25] Huang, W.-L., Hung, C.-Y., and Lin, I.-C., "Confidence-Based 6D Object Pose Estimation," *IEEE Transactions on Multimedia*, Vol. 24, 2022, pp. 3025–3035. https://doi.org/10.1109/TMM.2021.3092149

[26] Lee, T., "Bayesian Attitude Estimation with the Matrix Fisher Distribution on SO(3)," *IEEE Transactions on Automatic Control*, Vol. 63, No. 10, 2018, pp. 3377–3392. https://doi.org/10.1109/TAC.2018.2797162

[27] Loshchilov, I., and Hutter, F., "Decoupled Weight Decay Regularization," *7th International Conference on Learning Representations (ICLR)*, OpenReview.net, New Orleans, LA, 2019, https://openreview.net/forum?id=Bkg6RiCqY7.

[28] Zhang, Z., and Scaramuzza, D., "A Tutorial on Quantitative Trajectory Evaluation for Visual (-Inertial) Odometry," *2018 IEEE/RSJ International Conference on Intelligent Robots and Systems (IROS)*, Inst. of Electrical and Electronics Engineers, New York, 2018, pp. 7244–7251. https://doi.org/10.1109/IROS.2018.8593941

R. Windhorst
*Associate Editor*